\documentclass[letterpaper,journal]{IEEEtran}
\usepackage{amsmath,amsfonts}
\usepackage{array}
\usepackage{textcomp}
\usepackage{stfloats}
\usepackage{url}
\usepackage{verbatim}
\usepackage{graphicx}
\usepackage{cite}
\usepackage{xcolor}
\usepackage{subcaption}
\usepackage{mathtools}  
\usepackage{amssymb}
\usepackage{amsthm}
\usepackage{tabulary}
\usepackage{booktabs}
\usepackage[ruled,vlined]{algorithm2e}
\usepackage{algorithmic}
\usepackage{hyperref}
\usepackage{setspace}
\usepackage{subfiles}
\usepackage{multirow}
\usepackage{bm}
\allowdisplaybreaks

\newcommand{\RN}[1]{\uppercase\expandafter{\romannumeral#1\relax}}

\title{
  TOP: Trajectory Optimization via Parallel Optimization towards Constant Time Complexity

}

\author{
        \IEEEauthorblockN{
        Jiajun Yu\IEEEauthorrefmark{2}\textsuperscript{1, 2}, 
        Nanhe Chen\IEEEauthorrefmark{2}\textsuperscript{1, 2}, 
        Guodong Liu\textsuperscript{2, 3}, 
        Chao Xu\textsuperscript{1, 2},
        Fei Gao\textsuperscript{1, 2},
        and Yanjun Cao\textsuperscript{1, 2} 
        }
        \vspace{-1.0cm}
        }

\begin{document}

% \twocolumn[{
% \renewcommand\twocolumn[1][]{#1}
\maketitle
% \begin{center}
%   \setlength{\abovecaptionskip}{-0.10cm}
%   \setlength{\belowcaptionskip}{0.25cm}
%   \centering
%   \includegraphics[width=\textwidth,height=5cm]{Figures/yjj/test1.png}
%   \captionof{figure}{A quadrotor orbits a UGV by applying CoNi-MPC controller with a pre-computed circular trajectory in the UGV non-inertial frame. 
%           (a) is the accumulated shots of the quadrotor from the view of a camera on the UGV, which shows the relative circular trajectory of the quadrotor.
%           (b) shows the experiment from a third-person view in the world frame, in which the flight trajectory appears chaotic along with the UGV S-shape trajectory.}
%   \label{top_figure}
%   \end{center} 
  % }]

% \begingroup
% \renewcommand\thefootnote{\IEEEauthorrefmark{2} }
% \footnotetext{\textbf{Equal contribution}}
% \renewcommand\thefootnote{}
% \footnotetext{This work was supported by National Nature Science Foundation of China under Grant 62103368. (Corresponding authors: Yanjun Cao, Chao Xu.\tt\footnotesize \{Jiajun Yu, yanjunhi, cxu\}@zju.edu.cn)}
% \renewcommand\thefootnote{\textsuperscript{1} }
% \footnotetext{State Key Laboratory of Industrial Control Technology, Institute of Cyber-Systems and Control, Zhejiang University, Hangzhou, 310027, China.}
% \renewcommand\thefootnote{\textsuperscript{2} }
% \footnotetext{Huzhou Institute of Zhejiang University and Huzhou Key Laboratory of Autonomous Systems, Huzhou, 313000, China.}
% % \renewcommand\thefootnote{\textsuperscript{3} }
% % \footnotetext{Huzhou Key Laboratory of Autonomous Systems, Huzhou, 313000, China.}
\begingroup
\renewcommand\thefootnote{\IEEEauthorrefmark{2} }
\footnotetext{\textbf{Equal contribution}}
\renewcommand\thefootnote{}
\footnotetext{This work was supported by National Nature Science Foundation of China under Grant 62103368. (Corresponding authors: Yanjun Cao, Chao Xu.\tt\footnotesize \{yanjunhi, cxu\}@zju.edu.cn)}
\renewcommand\thefootnote{\textsuperscript{1} }
\footnotetext{State Key Laboratory of Industrial Control Technology, Institute of Cyber-Systems and Control, Zhejiang University, Hangzhou, 310027, China.}
\renewcommand\thefootnote{\textsuperscript{2} }
\footnotetext{Huzhou Institute of Zhejiang University and Huzhou Key Laboratory of Autonomous Systems, Huzhou, 313000, China.}
\renewcommand\thefootnote{\textsuperscript{3} }
\footnotetext{Tsinghua University, Beijing, 100084, China.}
\endgroup

\begin{abstract}

  Optimization has been widely used to generate smooth trajectories for motion planning.
  However, existing trajectory optimization methods show weakness when dealing with large-scale long trajectories.
  % important modules in motion planning. 
  % 再指出现有方法的不足
  % Existing methods usually optimize a parametrized polynomial trajectory for smoothness and efficiency. 
  % 分析现有方法存在不足的原因
  % However, the polynomial trajectory optimization problem introduces a large number of parameters when the trajectory needs to be precise and long. 
  % The state-of-the-art trajectory optimization algorithms only achieve linear time complexity, requiring over one second to optimize when the number of trajectory segments reaches seventy.
  % Thanks to the development of modern parallel computing architectures, many researchers has been working on the parallel optimization algorithm in the robotics field. 
  % 摆出需要解决的问题
  % How to leverage the parallel computing to solve the trajectory optimization problem efficiently is a still unsolved task.
  % How to leverage parallel computing to efficiently solve the trajectory optimization problem remains an unsolved task. 
  Recent advances in parallel computing have accelerated optimization in some fields, but how to efficiently solve trajectory optimization via parallelism remains an open question.
  % 提出解决方案
  In this paper, we propose a novel trajectory optimization framework based on the Consensus Alternating Direction Method of Multipliers (CADMM) algorithm, which decomposes the trajectory into multiple segments and solves the subproblems in parallel.
  % 分析解决方案为什么能解决提出得我问题
  The proposed framework reduces the time complexity to \(O(1)\) per iteration with respect to the number of segments, compared to \(O(N)\) of the state-of-the-art (SOTA) approaches.
  % 说明解决方案的优势
  % Furthermore, we integrate convex linear and quadratic constraints with closed-form solutions to speed up the optimization, and we also present numerical solutions for general inequality constraints. 
  Furthermore, we introduce a closed-form solution that integrates convex linear and quadratic constraints to speed up the optimization, and we also present a numerical solution for general convex inequality constraints.
  % 实验
  A series of simulations and experiments demonstrate that our approach outperforms the SOTA approach in terms of efficiency and smoothness.
  Especially for a large-scale trajectory, with one hundred segments, achieving over a tenfold speedup.
  %Specifically, it achieves a 10 times speedup once the number of trajectory segments reaches 60.
  To fully explore the potential of our algorithm on modern parallel computing architectures, we deploy our framework on a GPU and show high performance with thousands of segments.
\end{abstract}


\vspace{-0.20cm}
\section{Introduction}

% \IEEEPARstart{R}{ecently}, quadrotors or drones, due to their agility and
% lightweight nature, have been widely used in surveillance, search-and-rescue,
% and cinematography. The ability to autonomously plan and execute trajectories 
% is a key feature for these systems. The trajectory optimization problem for multi-rotor 
% aircraft is a challenging task due to the high degree of freedom and the non-linear
% dynamics of the system. Currently, collocation-based optimization
% methods, such as TrajOpt [1], solve a non-linear program(NLP) using non-linear 
% optimization algorithms like sequen-tial quadratic programming (SQP) [2]. 
% The structure of the NLP leads to at least \( O(N^2) \) time complexity concerning
% the number of waypoints to satisfy second-order optimality criteria. Therefore, 
% increasing the density of waypoints quadratically increases the computation time.
% The art-of-the-state trajectory optimization algorithms for multicopters, 
% such as the gcopter algorithm, has been widely used in the industry, 
% which already reduces the complexity of trajectory optimization 
% from \( O(N^2) \) to \( O(N) \). However, the complexity of the trajectory optimization 
% algorithm is still high, especially for long horizon tasks.

\IEEEPARstart{T}{rajectory} optimization is an essential part of the motion planning for the mobile robots. 
It has been extensively studied in the control community for general robotics systems\cite{controlCommunity}. 
Its primary purpose is to refine a path generated by path planning into a time-parameterized trajectory that satisfies various constraints and performance goals. 
Thanks to the great advantage of differential flatness\cite{Flatness}, we can recover all the states and inputs of a flat system from its flat output with finite derivatives. 
%add some ref?
This property is widely applied in the trajectory optimization for mobile robots\cite{kumarMinjerk}\cite{FlatnessBased}, making the problem tractable. 
However, the traditional trajectory optimization methods still suffer from high time complexity, especially for large-scale tasks. 
TrajOpt\cite{trajOpt}, solves a non-linear programming (NLP) problem using non-linear optimization algorithms like sequential quadratic programming (SQP). 
% The structure of NLP has a time complexity of at least \( O(N^2) \).
Due to its problem structure, the computational complexity of this NLP is at least \( O(N^2) \). 
Even the state-of-the-art (SOTA) algorithm, GCOPTER\cite{Gcopter}, which achieves \( O(N) \) time complexity operations on this representation under various planning requirements, still underperforms on large-scale problems. 
Here, \(N\) denotes the number of pieces, referred to as segments in this paper.

\begin{figure}[!t]
  \centering
  \includegraphics[width=0.476\textwidth]{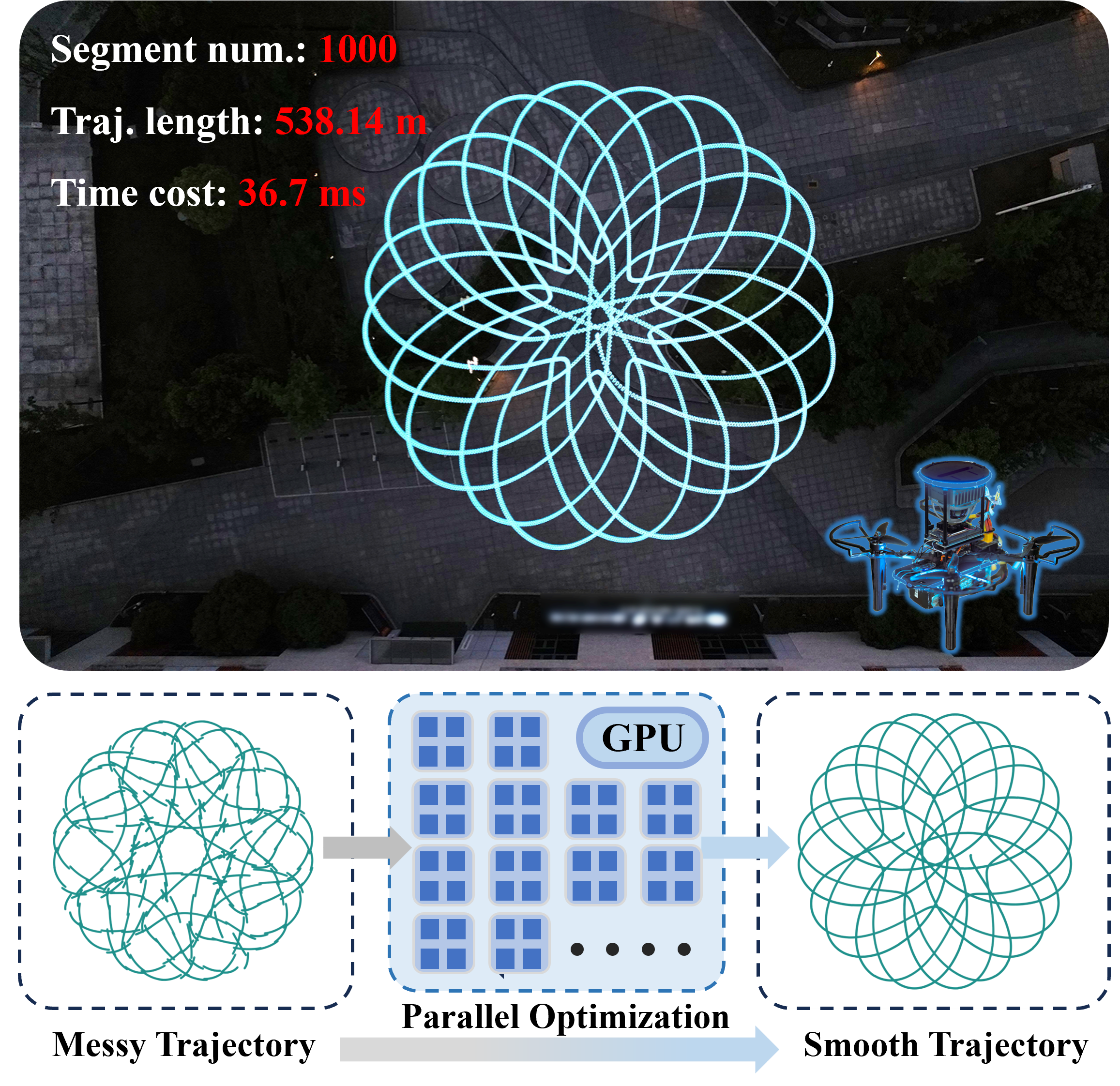}
  \caption{A real-world experiment of a quadrotor executing a large-scale trajectory generated by the proposed method. Top: A quadrotor executes the large-scale trajectory and the metrics. Bottom: The parallel optimization process of the trajectory is achieved via GPU.}
  \label{fig:real_gpu_test}
  \vspace{-6mm}
\end{figure}
% \begin{table}[h!]
%   \centering
%   \begin{tabular}{c|c|c|c|c|c}
%   \hline
%   \multirow{3}{*}{First } & Second  & Third  & Fourth  & Fifth  & Sixth  \\
%   \cline{2-6}
%    & Second  & Third  & Fourth  & Fifth  & Sixth  \\
%   \cline{2-6}
%    & Second  & Third  & Fourth  & Fifth  & Sixth  \\
%   \hline
%   \end{tabular}
%   \caption{A Table with Six Columns and First  Spanning Three Rows}
%   \end{table}

Inspired by the development of modern parallel computing architectures, many researchers have been working on the parallel optimization algorithms in the robotics field. 
% Due to the distributed property, prevailing works mainly focus on multi-agent scenarios, such as Collaborative Simultaneous Localization and Mapping (CSLAM), multi-robot task assignment, and collaborative trajectory optimization, etc. 
% However, the robots communicate with neighboring units over a bandwidth-constrained network, introducing communication delays, which makes distributed optimization unsuitable for real-time planning and control.
% In contrast, single-robot settings do not face these problems. 
Wang et al.\cite{trajectorySplitting} achieve a parallel structure by splitting the manipulator's joint angle trajectory optimization into a sequence of convex quadratic programming problems (QPs) using the Consensus Alternating Direction Method of Multipliers (CADMM) algorithm\cite{Boyd}. 
However, the safety constraints lack convex guarantees, which may result in the ADMM algorithm failing to converge.
% However, the safety constraints mentioned is non-convex, result in lacking of convergence guarantees. 
Furthermore, its low-order continuity makes it unsuitable for high-performance mobile robots requiring smooth motion, such as agile motion control for multicopters.

To enhance the performance of parallel trajectory optimization, we propose a novel framework named \textbf{TOP}.
% capable of solving the consensus problem in parallel. 
% We use it to reformulate the trajectory optimization problem.  
The foundation of \textbf{TOP} lies in the optimality conditions\cite{Gcopter} for minimizing control effort and the CADMM algorithm framework for high-order continuity. 
% The former guarantees the existence of the optimal solution, and the latter ensures the convergence of the algorithm. 
Specifically, we reformulate the trajectory optimization problem in a piecewise manner, where the optimization of each segment is formulated as a QP problem.
The high-order continuity between segments is satisfied by a set of corresponding consensus variables. 
% see Fig.\ref{fig:consensus_optimization}. 
In this way, our approach reduces the time complexity of trajectory optimization from $O(N)$ to $O(1)$ per iteration. 
Then we address the convex linear and quadratic inequality constraints using the indicator functions and projection methods for a closed-form solution. 
Furthermore, we also provide a numerical solution for readers to handle the general convex nonlinear constraints. 
Finally, we conduct a series of simulations and experiments to demonstrate that our approach outperforms the SOTA approach in terms of efficiency and smoothness.
Particularly for a large-scale trajectory with one hundred segments, achieving a speedup of over ten times.
To fully explore the potential of our algorithms on modern parallel computing architectures, we deploy our framework on a GPU and show high performance with thousands of segments.
As shown in Fig.\ref{fig:real_gpu_test}, a real-world experiment of a quadrotor flying a single large-scale trajectory of over 500 meters with 1000 segments, where the optimization time is less than one second. 
% What's more, the CADMM
% algorithm is used to solve the subproblems, which is known for its
% feature of fast convergence to modest accuracy—sufficient for many 
% applications—within a few tens of iterations.And the UAV trajectory
% Planning does not require high accuracy, even not to the local minimum,
% so the CADMM algorithm is suitable for this task.

% \begin{itemize}
%         \item Accurate state estimation of the agent and/or the target in the world frame;
%         \item Prior knowledge of the target's motion model for accurate prediction;
%         \item Continuous trajectory re-planning of the agent to be responsive and adaptive to the target's motion.
% \end{itemize} 
% }

% {\color{red} To the best of our knowledge, this is the first work realizing complex interaction between a drone and a target without state estimation in the world frame.
% Our work has made the following contributions: 
% \begin{itemize}
%         \item A novel and systematic framework for drone-target interaction based on  relative estimation;
%         \item A non-linear model predictive control formulation within a non-inertial target frame and extensive simulations to testify its feasibility;
%         \item Various real-world experiments conducted to show the potential application scenarios, including leader-following, aggressive directional landing, dynamic rings crossing, and orbit flight.

% \end{itemize}
% }

The contributions are summarized as follows:
\begin{itemize}
\item We reformulate the trajectory optimization in a parallel optimization framework, which reduces the time complexity from $O(N)$ to $O(1)$  while maintaining high-order continuity. 
%\item We propose two approaches to address the inequality constraints: one using indicator functions and projection methods for a closed-form solution, and the other approach is to transform all convex nonlinear constraints into a unified solving format, and employ slover to address each subproblem, thus yielding a numerical solution.
\item We propose two approaches to address the inequality constraints: one for convex linear and quadratic inequality constraints with a closed-form solution to speed up the optimization, and the other for general convex inequality constraints with a numerical solution.
\item A series of simulations and experiments demonstrate that our approach outperforms the SOTA method in efficiency and smoothness. 
Especially for a large-scale trajectory, with one hundred segments, achieving over a tenfold speedup. 
% \textcolor{red}{The framework is also deployed on a GPU, fully realizing the potential of our algorithms on modern parallel computing architectures.}
The framework is also deployed on GPU to explore the potential of our algorithm on modern parallel computing architectures.
\end{itemize}

\vspace{-0.30cm}
\section{Related Work}
\subsection{Motion Planning}

Motion planning algorithms are generally categorized into graph search-based methods\cite{SearchBase}, sampling-based methods\cite{SampleBase}, and trajectory optimization methods. 
Graph search-based methods, such as Dijkstra\cite{D*A*}, A*\cite{D*A*}, and JPS\cite{Jps}, discretize the configuration space into a graph and use search algorithms to find the optimal path. They are effective in low-dimensional space but suffer from the curse of dimensionality.
Sampling-based methods, such as PRM\cite{PRM}, RRT\cite{RRT}, and RRT*\cite{RRT*}, construct trees or graphs via random sampling in continuous space, enabling rapid exploration of feasible paths. 
They are scalable to high-dimensional problems but may take unaffordable time cost to get optimal solutions in real-time. 
Trajectory optimization methods, such as TrajOpt\cite{trajOpt}, GPOPS-II\cite{GPOPS-II}, and GCOPTER\cite{Gcopter}, numerically optimize trajectories in continuous space to balance various constraints and control goals (e.g., minimal time or minimal energy). They are designed for high-quality applications but suffer from high time complexity. 
For example, TrajOpt, solves an NLP using non-linear optimization algorithms like SQP. 
The underlying NLP structure exhibits at least \( O(N^2) \) time complexity.
The collocation method used in GPOPS-II has a time complexity of \( O(N^3) \) with respect to the number of collocation points, due to solving large systems of equations for each collocation point.
The GCOPTER, reduces the complexity of trajectory optimization to \( O(N) \) per iteration by utilizing a novel trajectory representation based on optimality conditions\cite{Gcopter}.
However, the complexity still grows linearly with the number of segments, making it unsuitable for large-scale tasks.
\vspace{-0.2cm}
\subsection{Parallel Optimization}

Recently, parallel optimization algorithms have attracted great attention in the robotics field. 
% The distributed property of parallel optimization algorithms let it be suitable for multi-robotsystems, such as Collaborative Simultaneous Localization and Mapping (CSLAM)\cite{D2slam} and collaborative trajectory optimization\cite{distributedMotionPlanning}, etc. 
% Shorinwa et al.\cite{distributedMultiRobot1}\cite{distributedMultiRobot2  } present a comprehensive survey of existing distributed optimization algorithms and classify them into three main categories. 
Existing methods mainly focus on multi-agent scenarios\cite{D2slam}, with limited research applying to single-agent trajectory optimization.
% \textcolor{red}{However, robots exchange information with its neighbors over a bandwidth-limited network with communication delays, which renders the inapplicable of the distributed optimization for real time planning and control.
% In contrast, single-robot settings do not face these communication bottlenecks or risks of link failure. }
Singh et al.\cite{ADMM_QP} propose a distributed framework by decomposing the mobile-manipulator trajectory optimization into a sequence of convex QPs. 
However, the concept of splitting the trajectory into multiple segments is not addressed in this work.
Wang et al.\cite{trajectorySplitting} achieve a parallel structure by splitting the manipulator's joint angle trajectory into multiple segments, and using CADMM to guarantee the continuity between the segments. 
However, the running speed of the algorithm framework has not been fully explored. 
Furthermore, its low-order continuity makes it unsuitable for high-quality control for multicopters. 
Leu et al.\cite{longHorizon} consider the scenarios requiring large-scale motion planning but still provide a nonsmooth trajectory.
% and note that existing methods suffer from naive initialization\cite{No_cadmm_naive}.
%However, the shortcomings mentioned above still exist, which introduces the need for further research.
% \todo{find two more refs. two is too little.}
In this paper, we reformulate the trajectory optimization problem into a parallel framework while enforcing the high-order continuity between the segmented trajectory pieces. 
% By introducing consensus variables, we ensure high-order continuity between the segmented trajectory pieces.
% Meanwhile, linear and quadratic constraints, as well as general constraints, are considered in the algorithm framework.
Our algorithm reduces the time complexity of trajectory optimization from \( O(N) \) to \( O(1) \) in a single iteration. 
%And it integrates convex linear and quadratic constraints into the algorithm framework, creating a cohesive system.
% To overcome these shortcomings of the above methods, we present a novel framework based on the CADMM algorithm, decomposing the trajectory into segments and solving the subproblems in parallel. It not only reformulates the fundamental minimum control problem but also integrates the convex equality and inequality constraints into a single iterative process. Additionally, we provide both a closed-form solution and an optimizer for practitioners to choose from.

\vspace{-0.20cm}
% \section{Problem Formulation And \\
% ~~~~Mathematical Background}
% \section{TOP:Trajectory Optimization Parallelly}
\section{Parallel Trajectory Optimization Algorithm}

In this section, we present our parallel trajectory optimization algorithm \textbf{TOP} based on the CADMM. %addressing both convex linear and nonlinear constraints. 
First, we reconstruct the constrained minimum control effort problem in a piecewise manner (Sec. \ref{sec:preliminaries}). 
Then, we reformulate the origin problem into a parallel consensus problem by adding consensus variables to ensure the continuity of the trajectory (Sec. \ref{sec:problem_formulation}).
%The added consensus variables of the new problem are used to ensure the continuity of the trajectory, which is a key feature of our method, see Fig.\ref{fig:consensus_optimization}.
Next, we propose two approaches to address both convex linear and nonlinear constraints: one is a closed-form solution, and the other is a numerical solution (Sec. \ref{sec:inequality_constraints}). 
Finally, we present the implementation details of our method (Sec. \ref{sec:implementation}).
\vspace{-0.3cm}
\subsection{Preliminaries}
\label{sec:preliminaries}
By leveraging the property of differential flatness, we can generate an optimal trajectory for the dynamic's flat output $\sigma(t) : \mathbb{R} \mapsto \mathbb{R}^m, t \in [0, T]$, which enables us to know the full states and inputs of the dynamic system.

The trajectory optimization problem can be formulated as a minimum control effort problem, i.e.,
\begin{subequations}  \label{minimum_control_effort}
  \begin{align}
    \min_{\sigma{(t)}}~& \int_0^T \sigma^{(p)}(t)^\top \mathbf{W} \sigma^{(p)}(t)dt, \\
    s.t.~& \sigma^{[p-1]}(0) = \overline{\sigma}_0,~ \sigma^{[p-1]}(T)=\overline{\sigma}_N, \label{initial_and_final}\\
    & ~ \mathcal{G}(\sigma(t),\dot{\sigma}(t),\ldots,\sigma^{(p)}(t))\preceq\mathbf{0},~\forall t \in [0,T],\label{general_constraints}
  \end{align}
\end{subequations}
% where the chain of \(n\)-integrators of the flatness output is denoted as \(\sigma^{[n]} = [\sigma^{\top}, \dot{\sigma}^{\top}, \ldots, {\sigma^{(n-1)}}^{\top}]\).  $\mathbf{W} \in \mathbb{R}^{m \times m}$ is a diagonal matrix scales the cost among different dimensions, $\overline{\sigma}_0\in \mathbb{R}^{ms}$ the initial condition, $\overline{\sigma}_N\in \mathbb{R}^{ms}$ the terminal condition, $\sigma^{(s)}$ the control input, and $\mathcal{G}$ consists of equivalent constraints.
% where the chain of \(n\)-integrators of the flatness output is denoted as \(\sigma^{[n]} = [\sigma^{\top}, \dot{\sigma}^{\top}, \ldots, {\sigma^{(n-1)^{\top}}}]^\top \). Here, \(\mathbf{W} \in \mathbb{R}^{m \times m}\) is a diagonal matrix that scales the cost across different dimensions, $\overline{\sigma}_0\in \mathbb{R}^{ms}$ the initial condition, $\overline{\sigma}_N\in \mathbb{R}^{ms}$ the terminal condition, $\sigma^{(s)}$ the control input, and $\mathcal{G}$ consists of equivalent constraints. 
where the chain of \(n\)-integrators of the flat output is denoted as \(\sigma^{[n]} = (\sigma^{\top}, \dot{\sigma}^{\top}, \ldots, {\sigma^{(n)^{\top}}})^\top \). Here, \(\mathbf{W} \in \mathbb{R}^{m \times m}\) is a diagonal matrix that scales the cost across different dimensions, \(\overline{\sigma}_0 \in \mathbb{R}^{mp}\) represents the initial condition, \(\overline{\sigma}_N \in \mathbb{R}^{mp}\) represents the terminal condition, \(\sigma^{(p)}\) denotes the control input and \(\mathcal{G}\) consists of equivalent constraints.

% G could be any constraints related to $c$, privided by specified tasks. 
% \textit{Michiel et al.} describes \textit{Differentially flat system} in \cite{??}, whose feasible state space and inputs can be determined by serveral carefully selected flat outputs.
% The property of \textit{Differentially flat system}\cite{??}, separating the nonlinear tracking problem into real-time trajectory generation.

Given a set of time durations $\mathcal{T} = \{T_1,T_2,\dots,T_N\}$, where $T_i\in \mathbb{R}_{>0}$ and $N$ is the number of segments, we reformulate the problem in a piecewise manner:
\begin{subequations}
  \begin{align}
    \min_{\sigma(t)}~ & \sum_{i=1}^{N} (\int_{0}^{T_i} \sigma_i^{(p)}(t)^\top \mathbf{W} \sigma_i^{(p)}(t) dt),\\
    % s.t.~ &  \sigma_j^{[d-1]}(T_i)= \sigma_{j+1}^{[d-1]}(0)
    % ,~j \in \{1,\dots,N-1 \}  \label{continously_differentiable_constraints}\\
    s.t.~ &  \sigma_i^{[d-1]}(T_i)= \sigma_{i+1}^{[d-1]}(0)
    ,~i \neq N,  \label{continously_differentiable_constraints}\\
    & \sigma_1^{[p-1]}(0) = \overline{\sigma}_0, ~\sigma_N^{[p-1]}(T_{N})=\overline{\sigma}_N, \label{start_points_constraints}\\
    & \mathcal{G}_i(\sigma_i(t),\dot{\sigma}_i(t),\ldots,\sigma_i^{(p)}(t)) \preceq\mathbf{0},~\forall t \in [0,T_i], \label{user_defined_constraints}
  \end{align}
\end{subequations}
where \(d\) is the number of derivatives, and Eq.~(\ref{continously_differentiable_constraints}) ensures the corresponding order of continuity.
% where (\ref{continously_differentiable_constraints}) enforces the $d$ order continuity of the trajectory between the segment. 
% Since the high-order continuity is not promised in piecewise polynomial by nature, we introduce extra constraints \ref{continously_differentiable_constraints} to ensure $\sigma(t)$ is $d - 1$ times continuously differentiable at $t_i$ for any $i \in \{1,2,\dots,N \} $.

% As widely adopted, we use polynomial to parameterized trajectories. The $i$-th segment of trajectories $\sigma_{i}(t)$ is a linear combination of the basis function $\beta(t) = (1, x, \dots, x^d)$ defined in the time interval $[t_{i-1},t_i]$, i.e.,
% \begin{equation} \label{polynomial traj class}
%   \begin{aligned}
%     \sigma_i(t) = \mathbf{c_i}^\top \cdot \beta(t - t_{i-1}),\quad t \in [t_{i-1},t_i]
%   \end{aligned}
% \end{equation} 
% where $c_i \in \mathbb{R}^{m \times (d+1)}$ are coefficients of the polynomial trajectory, and $d$ is the degree of the basis functions.

% All coefficients parameterized the trajectories constructing $ \mathbf{c} = [c_1^\top,c_2^\top,...,c_N^\top]^\top$. 

% In the  view of matrix operation, $J(c)$ can be rewritten to:
% where $Q$ is a block diagonal matrix, with $\{Q(T_1),Q(T_2),...,Q(T_N)\} \in \mathbb{R}^{m(d+1) \times m(d+1)}$ along the diagonal.
\vspace{-0.2cm}
\begin{figure}[!t]
  \centering
  \includegraphics[width=0.476\textwidth]{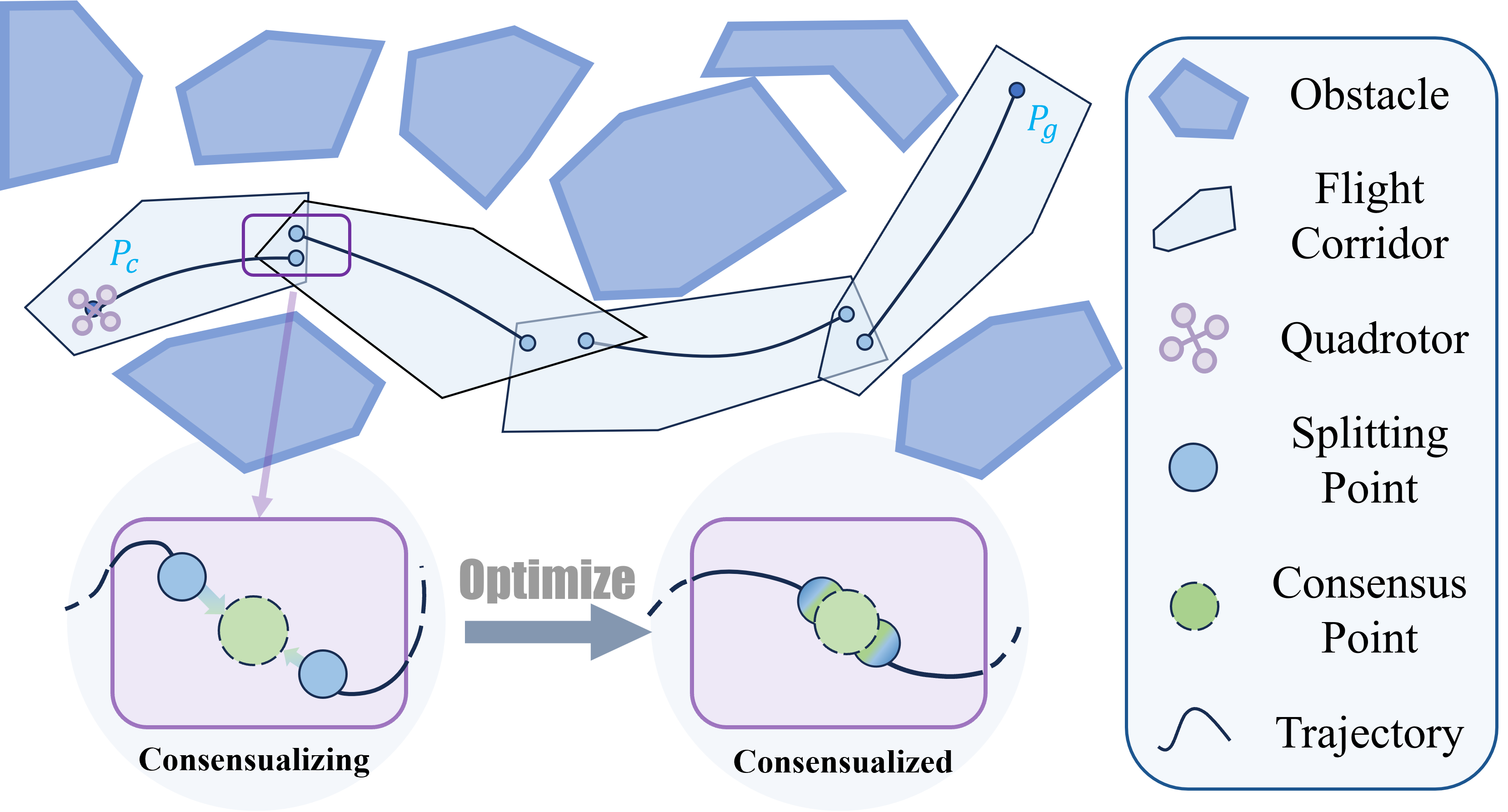}
  \caption{Consensus optimization segmented trajectory.}
  \label{fig:consensus_optimization}
  \vspace{-0.6cm}
\end{figure}

\begin{figure}[!b]
  \centering
  \vspace{-0.5cm}
  \includegraphics[width=0.476\textwidth]{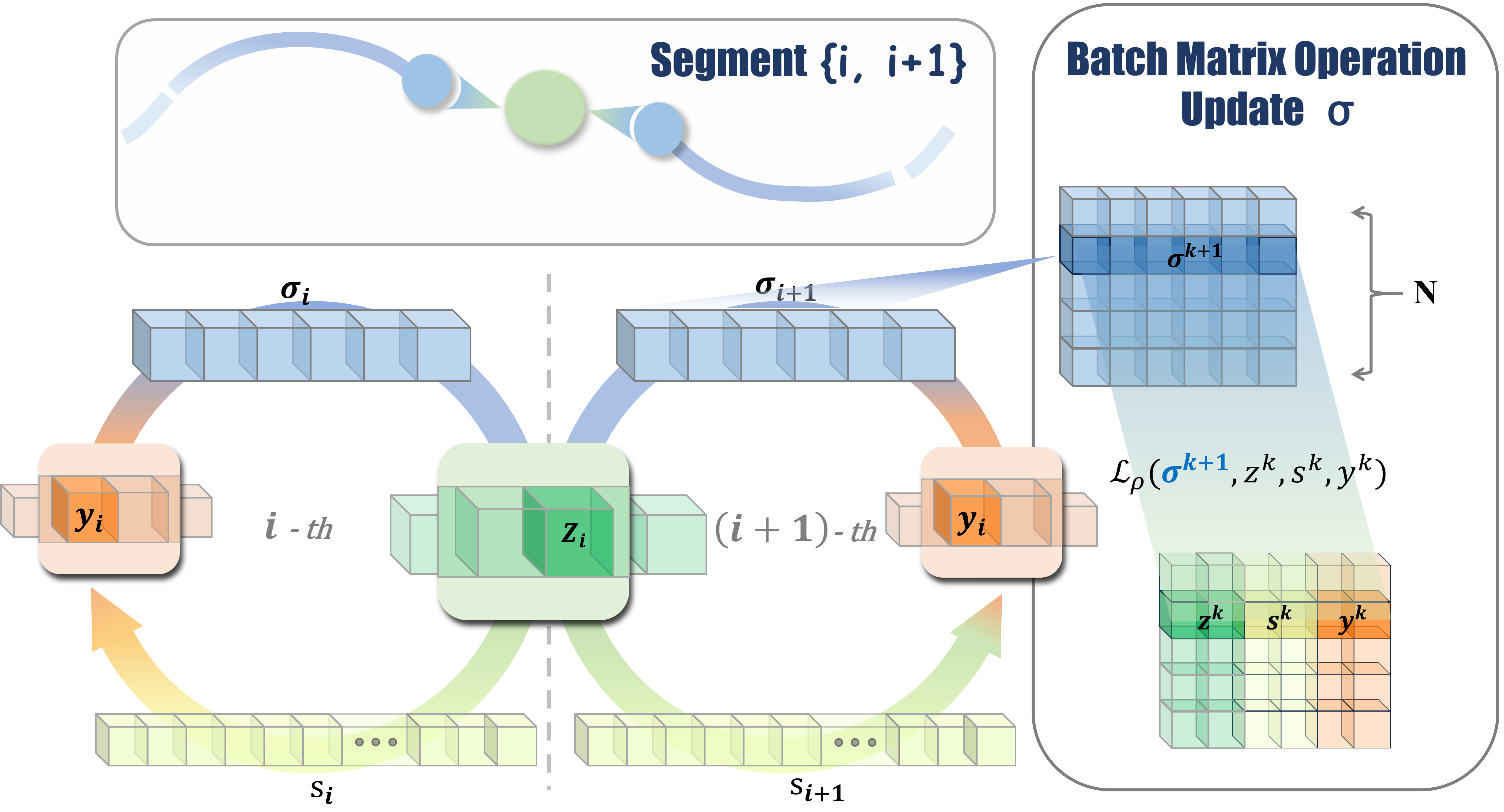}
  \caption{Parallel optimization at $(k+1)$-th iteration.}
  \label{fig:process of parallel optimization}
  % \vspace{-0.7cm}
\end{figure}

\vspace{-0.2cm}
% \subsection{Parallel Optimization}
\subsection{Problem Formulation} 
\label{sec:problem_formulation}
% \subsubsection{Reform to Consensus Problem}
\subsubsection{Reformulate as a consensus problem}
% The idea of distributed optimization is constructing homogeneous algorithm with each sub-problem sharing its intermediate decision variables with its one-hop neighbors at each iteration. In this section, we would tell why piecewise polynomial based trajectories are suitable to solve in a distributed fashion.

% To optimize the complete problem from the class of trajectories parameterized by piecewise polynomial, We may decompose the minimum control effort problem into sub-problems based on following considering: \\

% To reduce computing time complexity of the original problem with $Nm \times (d+1)$ variables. 
% A worthy view is to consider it being a consensus problem. 
% We conclude following properties of the complete problem to solve it in a parallel fashion:

% Considering a general consensus problem formulation:
Here we consider a general consensus problem formulation, with local variables $x_i \in \mathbb{R}^{n_i}, i \in \{1,\ldots,N\}$ and a global consensus variable $z \in \mathbb{R}^{n}$:
\begin{subequations}
\begin{align}
\min & ~\sum_{i=1}^N f_i(x_i), \\
% s.t. & ~ x_i - q_i = 0, ~ i = \{ 1, \ldots, N \}, \label{consensus constraints}
s.t. & ~ h(x_i)  = z, \label{consensus constraints}
\end{align}
\end{subequations}
% in which we have local variable $x_i \in \mathbb{R}^{n_i}, i \in \{1,\ldots,N\}$, and global variable $z \in \mathbb{R}^n$.
% (\ref{consensus constraints}) is the consensus constraint, where $(\widetilde{z_i})_j = z_{\mathcal{M}(i,j)}$, $\mathcal{M}$ mapping the local variable component $h(x_i)_j$ corresponding to some global variable component.
where $h:\mathbb{R}^{n_i} \mapsto \mathbb{R}^{n}$ is the mapping from local variables to the global consensus variable.
The consensus constraints in Eq. (\ref{consensus constraints}) ensure that all \(x_i\) reach an agreement.

Motivated by this approach, we reformulate the original problem into a consensus problem by considering the following aspects:

\begin{itemize}
 \item \textbf{Independent Problem:}
  The objective function of each segment depends solely on its local variables, allowing it to be divided into independent local objectives.
  % i.e., 
  % \begin{equation}
  %   \begin{aligned}
  %     \mathcal{J}_i = & \int_{0}^{T_i} \mathbf{c_i}^\top\beta^{(s)}(t) \mathbf{W} \beta^{(s)}(t)^\top \mathbf{c_i} dt \\
  %   \end{aligned}
  % \end{equation}
  % Furthermore, the convex constraints in (\ref{user_defined_constraints}) do not introduce any nonlinear relationship among the trajectory segments. 
  Meanwhile, the absence of coupling constraints in Eq.~(\ref{user_defined_constraints}) between segments enables each segment to be solved independently.
  
  \item \textbf{Consensus Constraints:}
  By introducing consensus variables, continuity constraints in Eq. (\ref{continously_differentiable_constraints}) are transformed into consensus constraints as described in Eq.(\ref{consensus constraints}).

\end{itemize}

Based on the previous analysis, 
% we introduce the global variables $\mathcal{Z} = \{{z_0}, \ldots z_N\}$, $z_i \in \mathbb{R}^{md}$, which define the positions and their \(d-1\) derivatives at the split points.
the original problem can be reformed into the consensus problem, i.e.,
% By introducing the consensus variables, we succeed in decomposing the original problem into a series of sub constrainted quadratic programming.
% Traditional methods solve the above multi-segment trajectory optimization problem jointly. We consider to solve the problem in a decentralized fashion. Consensus Alternating Direction Multipliers Method(C-ADMM) is able to solve the consensus problem parallelly. We use it to reformulate the (\ref{original_problem_formulation}) problem. In our new problem formation based on ADMM, each sub-problem optimizes the coefficients of one piece of the trajectory, the capability of which, is $\frac{1}{N}$ of the original problem. The formulation is specified in the following. 
% For simplicity, we only consider a trajectory optimzation problem (Euclidean Space in $R^3$ at most case)  in one-dimensional case.
% Specifically, we decompose the problem into a series of subproblems, each of which is a two-point boundary value problem (TPBVP) with two waypoints. The splitted position and higher order continuity constraints are enforced by a series of consensus variables. The trajectory will become smoother as the consensus variables are updated iteratively. Moreover, the safety constraints and dynamic feasibility constraints are also integrated into the algorithm framework formulated as a convex closed-form problem, which will be introducted in the next section.
\begin{equation}
  \begin{aligned}
    \label{Consensus_problem}
    \min_{\sigma(t)} ~& \sum_{i=1}^{N} (\int_{0}^{T_i} \sigma_i^{(p)}(t)^\top \mathbf{W} \sigma_i^{(p)}(t) dt), \\
    % s.t. ~&  \sigma_i^{[d-1]}(T_i)= z_i,
    % ~ \sigma_{i}^{[d-1]}(0) = z_{i-1}, 
    % s.t. & [\sigma_i^{[d-1]}(T_i),\sigma_{i}^{[d-1]}(0)]^\top = [z_{i-1}, z_{i}]^\top
    s.t. ~& \widetilde{\sigma_i}^{[d-1]} = \widetilde{{z_i}},
  \end{aligned}
\end{equation}
% $z_i \in \mathbb{R}^{m(2s-1)}$ is the consensus variable related to constrainting the high-order derivatives. 
where \(\widetilde{\sigma_i}^{[d-1]} = (\sigma_i^{[d-1]}(0)^\top, \sigma_i^{[d-1]}(T_i)^\top)^\top\) represents the derivatives of the flat output at both boundaries. 
The consensus variable \(z_i \in \mathbb{R}^{md}\) includes the consensus point along with its derivatives up to the $(d-1)$-th order, while \(\widetilde{z_i} = (z_{i-1}^\top, z_i^\top)^\top\) represents the boundaries of the \(i\)-th segment.
For simplicity, only continuous constraints are considered in the Eq. (\ref{Consensus_problem}).

% To solve the consensus problem in ADMM algorithm, we denote the full sub-problem as:

% % Yield convergence without assuming the minimum control effect problem to be strictly convex.
% % Leveraging the dual ascent method to construct a unconstrained optimization problem, the original problem is equal to the following:

% \begin{equation}
%   \begin{aligned}
%     \min_{\mathbf{c_i}} & (\mathbf{c_i}^\top \mathbf{Q}(T_i) \mathbf{c_i} + \frac{\rho}{2} ||\widetilde{M} (T_i) \cdot \mathbf{c_i} - \widetilde{\mathbf{z_i}}||_2^2) \\
%     s.t. &\widetilde{M} (T_i) \cdot \mathbf{c_i} = \widetilde{\mathbf{z_i}} \\
%     & G_i(c) \preceq 0,
%   \end{aligned}
%   \label{Consensus_augmented_Lagrangian_problem}
% \end{equation}
% where $\widetilde{z_i} = [z_{i},z_{i+1}]^\top$ are the components of $z$ involving in $i$-th piece,
% $\widetilde{M}(t) = [{\beta^{[d-1]}(0)}^\top,{\beta^{[d-1]}(t)}^\top]^\top$ maps coefficients in $i$-th piece to high-order derivatives in both sides,
% $u_i \in \mathbb{R} ^ {2m(2s-1)} $ is the scaled dual variable of the consensus constraint, defined as $u_i = y_i/ \rho$, where $\rho$ is the penalty parameter. 

\vspace{0.1cm}
\subsubsection{CADMM Parallel Optimization} 
We solve the consensus problem in the context of the CADMM algorithm\cite{Boyd}. 
To apply the algorithm, we first give the augmented Lagrangian function for the subproblem of each segment \(i\):
\begin{equation}
  \begin{aligned}
    \hspace{-0.4em} \mathcal{L}_{i} = \int_{0}^{T_i} \sigma_i^{(p)}(t)^\top \mathbf{W} \sigma_i^{(p)}(t) dt + \frac{\rho}{2} ||\widetilde{\sigma}_i^{[d-1]} - \widetilde{z_i} + u_i||_2^2.
  \end{aligned}
  \label{Consensus_augmented_Lagrangian_problem}
\end{equation}
%  respectively, which remain fixed during the iteration. 
Let $u_i \in \mathbb{R}^{2md}$ be the scaled dual variable for continuity constraints, defined as $u_i = y_i / \rho$, where $\rho$ is the penalty parameter. So the overall function is given by $\mathcal{L}_{\rho} = \sum_{i=1}^{N} \mathcal{L}_{i}$.
% Based on our C-ADMM algorithm framework, we can solve the problem in a distributed manner. 

% In order to solve the subproblems which are QPs parallelly in subsequent iterations. Then the consensus variable $z_i$ will average the final state of the previous segment, $i$ th, and the initial state of the next segment, 
% $i+1$ th, which will continuously connect the gaps during the iterations with the Lagrange multiplier $u_i$ corresponding the consensus constraint. Significantly, the other constraints, such as the obstacles avoidance, dynamic feasibility, etc, can be integrated into the algorithm framework, which will be introducted in the next section.The update rule of the iteration of the CADMM algorithm is:

% Then we formulate aforementioned problem into ADMM form and solve it as below:
% The algorithm to solve a CADMM problem iteratively can be given as:
The CADMM method to solve the consensus problem iteratively can be given as:
\begin{subequations}\label{Consensus_update}
  \begin{align}
    {\sigma_i}^{k+1} &= \arg \min_{\sigma_i} \mathcal{L}_i(\sigma_i,z_i^k,u_i^k), \\
    z_i^{k+1} &= \frac{1}{2}(\sigma_i^{[d-1]^{k+1}}(T_i) + \sigma_{i+1}^{[d-1]^{k+1}}(0)) , i \notin \left\{ {0,N} \right\},  \\
    u_i^{k+1} &= u_i^k + \widetilde{\sigma}_i^{[d-1]^{k+1}} - \widetilde{z_i}^{k+1}, \label{Consensus_dual_update}
  \end{align}
\end{subequations}
where $z_0$ and $z_N$, initialized by $\overline{\sigma}_0$ and $\overline{\sigma}_N$ respectively, are kept fixed during the iterations.
% where the updating of $c_i$ have certainly closed-form solution, due to the sub-problem remain its quadratic function formation. 
Fig. \ref{fig:process of parallel optimization} illustrates how optimization variables $\sigma$, scaled dual variables $u$, and consensus variables $z$ participate in the algorithm during iterations (the slack variables $s$ will be introduced in the next section). All variables are organized into tensors to make the computing process parallel.
\vspace{-0.5cm}
\subsection{Convex Inequality Constraints}
\label{sec:inequality_constraints}
% In this section, we will introduce how to deal with a set of convex inequality constraints in proposed framework to generate task-specific trajectories. 
% First, we will introduce how to handle linear inequality and quadratic inequality in a closed-form way. 
% Then, more general convex inequality constraints will be solved in a numerical way. 
% At last, we will leverage above closed-form methods to generate a collision-free and dynamically feasible trajectory as an example.
\vspace{-0.1cm}
\subsubsection{Linear Inequality Constraints}
% In this paper, we use flight corridors to represent safety constraints, which can be formulated as multiple linear inequality constraints. Such linear inequalities are easily handled by the ADMM algorithm. 
% Similar to linear equality constraints, linear inequality constraints can also be handled with ease. 
% We tackle linear inequality constraints via transforming them into linear equality constraints by introducing slack variables.\todo{1}
To address the linear inequality constraints, we introduce slack variables, converting the inequality constraints into equality constraints, making the procedure similar to Eq. (\ref{Consensus_update}). 
In trajectory optimization, collision-free space is often represented as a sequence of convex overlapping polytopes, defined by linear inequality constraints.
% convex polytopes are widely used to represent obstacle-free space, which can be defined as a set of linear inequalities. 
% Liu et al.\cite{liuFlightCorridor} proposed a method to generate a collision-free trajectory based on the flight corridor.
% The flight corridor is a sequence of convex overlapping polytopes.\todo{2}
% Wang et al. \cite{Firi} introduced FIRI method, which generates high-quality flight corridor, a sequence of convex overlapping polytopes.
Hereafter, we take a simple linear inequality constraint as an example: 
\begin{equation} 
\begin{aligned}
\arg \min_x&~f(x), \\
s.t.&~\mathbf{A}x \preceq b,
\end{aligned}
\end{equation}
where \(\mathbf{A} \in \mathbb{R}^{m \times n}\), \(b \in \mathbb{R}^{m}\), and optimization variable \(x \in \mathbb{R}^{n}\). 
After introducing slack variable \(s \in \mathbb{R}^m\), the inequality constraint can be transformed into equality constraint:
\begin{equation}
\begin{aligned}
\arg \min_{x,s}&~f(x) + I_{s \geq 0}(s), \\
s.t.&~\mathbf{A} x + s = b,\\
\end{aligned}
\label{linear_inequality}
\end{equation}
where $I_{s \geq 0}(s)$ is the indicator function, formulated as:
\begin{equation}
I_{s \geq 0}(s)=\left\{\begin{array}{cc}
+\infty & s<0, \\
0 & s \geq 0.
\end{array}\right.
\end{equation}
Finally, the update rule is the same as that for linear equality constraint in Eq. (\ref{Consensus_update}), with the update rule for the slack variable \(s\) in the $(k+1)$-th iteration as:
\begin{equation}
  s^{k+1} = \max(0, -(\mathbf{A}x^{k+1} - b + v^k)),
\end{equation}
where \(v\) is the corresponding scaled dual variable.
% We can apply the same processing approach to the $i$th trajectory segment, subject to the safety constraints of the flight corridor. The update rule corresponding to the safety constraints is:
% \begin{subequations}
% \begin{align}
% c_i^{k+1} &= \arg \min_c f(c) + \frac{\rho}{2}||A_i^s c_i -b_i + s_i + v_i ||_2^2 ,\\
% s_i^{k+1} &= \min(0, -(A_i^s c_i^{k+1} -b_i + v_i^k)) ,\label{solveSlack}\\
% v_i^{k+1} &= v_i^k + A_i^s c_i^{k+1}+ s_i^{k+1} -b_i.
% \end{align}
% \end{subequations}
% Where \(f(c)\) is the objective function of (\ref{Consensus_augmented_Lagrangian_problem}) for simplicity. The set \(\mathcal{P}_{i}^{\mathcal{H}}=\{\,c_i \in \mathbb{R}^{6} \mid A_{i}^s c_i \preceq b_{i}\}\) is a bounded convex polytope described by its H-representation \cite{hPrensentation}, forming the safety flight corridor. The matrix \(A_i^s \in \mathbb{R}^{MK \times 6}\) and \(b_i \in \mathbb{R}^{MK}\) are defined accordingly, where \(M\) represents the number of sample points per segment and \(K\) represents the number of flight corridors per segment. The slack variable \(s_i \in \mathbb{R}^{MK}\) enforces the safety constraints, while \(v_i \in \mathbb{R}^{MK}\) is the scaled dual variable for those constraints in the \(i\)th segment.
% \vspace{-0.1cm}
\subsubsection{Quadratic Inequality Constraints}
% Besides the obstacle avoidance constraints formulated as linear inequalities, dynamic feasibility constraints are equally crucial for trajectory optimization. 
Besides the linear inequality constraints, convex quadratic inequality constraints are also widely used for trajectory optimization, such as dynamic feasibility, target tracking, and minimal trajectory length. 
%  which are often formulated as quadratic inequalities constraints.
In this work, we address these constraints using a projection method that provides a closed-form solution.
% Below, we take a convex quadratic inequality constraint as an example:
An example of a convex quadratic inequality constraint is provided below:
% Taking the maximal velocity of dynamic feasibility constraint as an example, the formulation is:
\begin{equation}
  \begin{aligned}
  \arg \min_x & ~f(x), \\
  s.t. &~||\mathbf{B} x||^2 \leq c,
  \end{aligned}\
  \label{original_quadratic_problem}
  \end{equation}
% \begin{equation}
% \begin{aligned}
%   \left\| A_i^v(T_{ij}) c_i \right\|^2 \leq v_{max}^2,\\
% \end{aligned}
% \end{equation}
where \(\mathbf{B} \in \mathbb{R}^{m \times n}\), \(c \in \mathbb{R}_{>0}\) and optimization variable \(x \in \mathbb{R}^{n}\). 
We introduce the $\phi  =\mathbf{B} x,~\phi \in \mathbb{R}^m$, and replace the quadratic inequality constraint with the indicator function:$I_{||\phi||^2 \leq c}(\phi)$. 
Then the original problem Eq. (\ref{original_quadratic_problem}) can be transformed into:
\begin{equation}
  \begin{aligned}
  \arg \min_x&~ f(x) + I_{||\phi||^2 \leq c}(\phi), \\
  s.t.&~\phi=\mathbf{B} x.
  \end{aligned}
  \label{Quadratic_inequality}
  \end{equation}
Therefore, the update rule is identical to that for linear equality constraint in Eq. (\ref{Consensus_update}). Notably, the update rule for \(\phi\) is:
% where $T_{ij}$ is the time of the $j$th intra-sample point of the $i$th segment, $A_i^v(T_{ij}) \in \mathbb{R}^{3 \times 6}$, and $v_{\max}$ is the maximal velocity.After updating \(c_i\) in each iteration of the CADMM algorithm, we project \(c_i\) onto the feasible set, which can be formulated as:
\begin{equation}
  \phi^{k+1} = arg \min_{\phi} (I_{||\phi||^2 \leq c}(\phi) + \frac{\rho}{2}||\mathbf{B} x^{k+1} - \phi + w^k||^2_2),
\end{equation}
where $w$ is the scaled dual variable.
The process is equivalent to projecting the vector $\mathbf{B} x + w$ onto the constraint set $\{\phi \in \mathbb{R}^m \mid ||\phi||^2 \leq c\}$, which can be solved in a closed-form way:
\begin{equation}
  \phi =
\left\{
\begin{aligned}
  &\sqrt{\frac{c}{\left\| \mathbf{B} x + w \right\|^2}}(\mathbf{B} x + w)
  && \text{if }  ||\mathbf{B} x+w||^2 > c, \\
  &~\mathbf{B} x+w
  && \text{otherwise.}
\end{aligned}
\right.
\end{equation}

\vspace{-0.2cm}
\subsubsection{General Inequality Constraints}
% Closed-form solutions are not always available for every convex inequality constraint, which may be necessary in certain cases. 
% In many cases, closed-form solutions are not available. 
Although the closed-form solution holds for most constraints in multicopter trajectory optimization, certain specialized ones may not be satisfied. 
Giesen et al.\cite{inequalityConstraints} provide a universal approach capable of directly handling arbitrary non-linear convex inequality constraints. 
However, its mathematical form limits the possibility of obtaining a closed-form solution. 
In this work, we employ the L-BFGS \cite{LBFGS} algorithm to solve the subproblems when updating \(\sigma_i\) during the iterations. 
Specifically, it replaces the inequality constraints with an equality form by 
\begin{equation}
  g(x)=\max \left\{0, g_{0}(x)\right\}^{2},
  \label{componentwise maximum}
\end{equation}
where \(g_{0}(x)\) is original inequality constraint. 
By using Eq.(\ref{componentwise maximum}), we transform the convex inequality constraints into convex equality constraints, differing from the approach of introducing slack variables.
% Note that the $g(x)$ is also a convex function if $g_{0}(x)$ is convex. Though, that the constraints $g(x)=0$ is no longer affine. 
%However, it is still can be solved effectively, which will be introducted soon.
% Thus, in the following, we consider the equality constrained problem form: 
The $g(x)$ inherits convexity from $g_0(x)$, but the equality constraint $g(x)=0$ is no longer linear.
In summary, we consider the equality-constrained problem form: 

% \todo{rewrite}
% \vspace{-0.3cm}
\begin{equation}
  \begin{aligned}
    \underset{x}\min &~
    f(x), \\
    s.t.&~ g(x)=0.
  \end{aligned}
  \label{prob:all inequality}
\end{equation}
The augmented Lagrangian function for Eq. (\ref{prob:all inequality}) is: 
\begin{equation}
\begin{aligned}
  \label{general_constraint_aug}
  L_{\rho}(x, y)&=  f(x) + \frac{\rho}{2} \|g(x)\|^{2} + y^{\top} g(x), \\ 
\end{aligned}
\end{equation}
where $y \in \mathbb{R}^{p}$ is the Lagrange multiplier. 
Then the dual variable $ y^{k} $ will be updated similarly to the dual variable of the equality constraint Eq. (\ref{Consensus_dual_update}) in the $(k+1)$-th iteration.

% The advantage of this method is that it does not introduce any additional variables or subproblems to the original problem, unlike the aforementioned method. 
% This method is benifited from no additional variables or steps during iteration are introduced to the original problem.
This method benefits from solving general convex inequality constraints while avoiding the introduction of additional variables and iteration substeps to the original problem.
% \textcolor{red}{However, it cannot be transformed into a closed-form solution due to the max function in the inequality constraints.}
% Therefore, the optimization solver like L-BFGS should be used to solve the subproblems, which will make the algorithm slower than the closed-form solution, as demonstrated in the experimental section. 
However, numerical solvers like L-BFGS slow down the running speed, which is detailed in the experiments part Sec. \ref{Sec: Ablation}. 
Above all, it's an alternative option for users to choose the one that best matches their requirements.

\vspace{-0.2cm}
\subsection{Implementation Details}
\label{sec:implementation}
% \vspace{-0.1cm}
\subsubsection{Problem Definition of Minimum-jerk Trajectory}  
In this section, we construct a minimum-jerk trajectory $(p=3)$ through the proposed closed-form solution as an example. 
This trajectory simultaneously enforces collision-free and dynamical feasibility constraints.
Meanwhile, we parameterize the trajectory with a set of polynomials.
The flat output of the $i$-th segment  $\sigma_{i}(t)$ is a linear combination of the basis function $\beta(t) = (1, t, \dots, t^d)^\top$, i.e.,
\begin{equation} \label{polynomial traj class}
  \begin{aligned}
    \sigma_i(t) = \mathbf{c_i}^\top \cdot \beta(t - t_{i-1}),\quad t \in [t_{i-1},t_i],
  \end{aligned}
\end{equation} 
where $\mathbf{c_i} \in \mathbb{R}^{(d+1) \times m}$ represents the coefficients of the $i$-th polynomial trajectory, and $d$ (set to 5) is the degree of the basis function. 
% In the quadratic field, the feasible region is typically represented by the flight corridor. 
% Liu et al.\cite{liuFlightCorridor} proposed a method to generate a collision-free trajectory based on the flight corridor.
% The flight corridor is a sequence of convex overlapping polytopes.
% Wang Et al.\cite{Firi} introducted the FIRI, which gets more higher quality polytopes.
As for the collision-free constraints, we use the FIRI\cite{Firi} to generate the flight corridor, which can be formulated as multiple linear inequality constraints, addressed by Eq. (\ref{linear_inequality}). 
% We then apply a projection method to enforce dynamic feasibility constraints. 
% As an illustrative example, we focus on the constraint of maximum velocity. 
We then apply the projection method in Eq. (\ref{Quadratic_inequality}) to enforce dynamic feasibility constraints, using maximum velocity as an example.
For the convenience of closed-form operation, we squeeze the $\mathbf{c_i}$ matrix into a vector $c_i \in \mathbb{R}^{m(d+1)}$.
Considering the \(i\)-th segment (\(m = 3\)), the consensus subproblem is formulated as:
\begin{equation}
  \begin{aligned}
  % \arg \min_{c_i}& f(c_i), \\
  \arg \min_{c_i}~& c_i^\top \widetilde{\mathbf{Q}}(T_i) c_i, \\
  s.t. 
  % ~&c_i^\top \beta^{[d-1]}(0) = z_{i-1}, \\
  % ~&c_i^\top \beta^{[d-1]}(T_i)= z_i,\\ 
  ~& \widetilde{\mathbf{M}}(T_i)c_i = \widetilde{z_i},~ \mathbf{A}_i^o c_i \preceq b_{i}, \\
  & \left\| \mathbf{A}^v(T_{ij}) c_i \right\|^2 \leq v_{max}^2,
  \end{aligned}
\end{equation}
% Where \(f(c_i)\) is the objective function of (\ref{Consensus_augmented_Lagrangian_problem}) for simplicity. 
% where $\mathbf{Q} = \operatorname{diag}(\mathbf{Q_1}, \mathbf{Q_2}, \mathbf{Q_3}) \in \mathbb{R}^{18\times 18}, \mathbf{Q}_1 = \mathbf{Q}_2 = \mathbf{Q}_3 = \beta^{(s)}(t) \mathbf{W} \beta^{(s)}(t)^\top \in \mathbb{R}^{6\times 6}$ ,
% where $\widetilde{\mathbf{Q}} = \mathbf{I}_{m\times m} \otimes \mathbf{Q} $, $\mathbf{Q} = \beta^{(s)}(t) \mathbf{W} \beta^{(s)}(t)^\top \in \mathbb{R}^{(d+1)\times (d+1)}$, is a semi-positive matrix. 
where $\widetilde{\mathbf{Q}}(t) = \mathbf{I}_{m\times m} \otimes \mathbf{Q}(t) $. 
$\mathbf{Q}(t) = \int_{0}^{t} \beta^{(p)}(\tau) \beta^{(p)}(\tau)^\top d{\tau} \in \mathbb{R}^{(d+1)\times (d+1)}$ is a positive semidefinite matrix.
% The matrix $\widetilde{\mathbf{M}} = \mathbf{I}_{m\times m} \otimes \mathbf{M} $, $\mathbf{M} =[\beta^{0}(0)^\top, \dots, \beta^{0}(0)^\top, \beta^{(d-1)}(t)^\top, \dots ,\beta^{[d-1]}(t)^\top]^\top \in \mathbb{R}^{2d \times (d+1)}$, maps the coefficients in the $i$-th segment to high-order derivatives at both boundaries.
The matrix $\widetilde{\mathbf{M}}(T_i) = \mathbf{I}_{m\times m} \otimes \mathbf{M}(T_i)$ maps the coefficients to the high-order derivatives at both boundaries of the $i$-th segment, where $\mathbf{M}(t) =(\beta^{0}(0), \dots, \beta^{(d-1)}(0), \beta^{0}(t), \dots, \beta^{(d-1)}(t))^\top \in \mathbb{R}^{2d \times (d+1)}$.
% The matrix $\widetilde{\mathbf{M}}(t) = \mathbf{I}_{m\times m} \otimes (\beta^{[d-1]}(0),\beta^{[d-1]}(t))^\top$, where $\mathbf{M}(t) =\mathbf{I}_{m\times m} \otimes \beta^{[d-1]}(t) \in \mathbb{R}^{d \times (d+1)}$ maps the coefficients of the $i$-th segment to the high-order derivatives of the boundaries when $t\in\{0,T_i\}$.
% The set \(\mathcal{P}_{i}^{\mathcal{H}}=\{\,c_i \in \mathbb{R}^{6} \mid \mathbf{A}_i^s c_i \preceq b_{i}\}\) described by its H-representation \cite{hPrensentation}, forming the safety flight corridor. 
The free space is encoded as a convex polytope \(\mathcal{P}_{i}^{\mathcal{H}}=\{\,c_i \in \mathbb{R}^{m(d+1)} \mid \mathbf{A}_i^o c_i \preceq b_{i}\}\), which is part of the safety flight corridor. 
The matrix \(\mathbf{A}_i^o \in \mathbb{R}^{MK \times m(d+1)}\) and \(b_i \in \mathbb{R}^{MK}\), where \(M\) represents the number of sample points per segment and \(K\) represents the number of hyperplanes per polytope. 
% The slack variable \(s_i \in \mathbb{R}^{MK}\) and the scaled dual variable \(v_i \in \mathbb{R}^{MK}\) are associated with the \(i\)-th segment.
As for the dynamic constraints, $T_{ij}$ is the timestamp of the $j$-th intra-sample point of the $i$-th segment, $j\in\{1,2,\dots,M\}$, $\mathbf{A}^v(T_{ij}) \in \mathbb{R}^{m \times m(d+1)}$, and $v_{\max}$ is the maximal velocity. 
Finally, the update rule of the $(k+1)$-th iteration is:
\begin{subequations}
\begin{flalign}
  % \!& c_i^{k+1} = \arg \min_{c_i}~ \mathbf{c_i}^\top \mathbf{Q}(T_i) \mathbf{c_i} 
  % + \frac{\rho}{2} \left\| \widetilde{\mathbf{M}}(T_i) \cdot \mathbf{c_i} - \widetilde{\mathbf{z_i}} + \mathbf{u_i} \right\|_2^2 \notag \\
  % \!& \qquad + \frac{\rho}{2} \left\| A_i^s c_i - b_i + s_i^k + v_i^k \right\|_2^2 
  % + \sum_{j=1}^{M\mathbf{I}_{m\times m} \otimes \mathbf{Q}(t)} \frac{\rho}{2} \left\| A^v(T_{ij}) c_i - d_{ij}^k + w_{ij}^k \right\|_2^2, \label{QpSubProblem} \\[1ex]
  %
  % \!& c_i^{k+1} = \left[ 2\mathbf{Q}(T_i) + \rho \widetilde{\mathbf{M}}(T_i)^\top \widetilde{\mathbf{M}}(T_i) 
  % + \rho (A_i^s)^\top A_i^s + \right. \notag \\
  % \!& \left. \qquad \rho \sum_{j=1}^{M} A^v(T_{ij})^\top A^v(T_{ij}) \right]^{-1} \cdot \left[ \widetilde{\mathbf{M}}(T_i)^\top(\widetilde{\mathbf{z_i}} - \mathbf{u_i}) 
  % + (A_i^s)^\top(b_i - s_i^k - v_i^k) 
  % + \sum_{j=1}^{M} A^v(T_{ij})^\top(d_{ij}^k - w_{ij}^k) \right], \\[1ex]
  \!& \hspace{-0.5em} c_i^{k+1} = \Big[ 2 \widetilde{\mathbf{Q}}(T_i) 
  + \rho\, \widetilde{\mathbf{M}}(T_i)^\top \widetilde{\mathbf{M}}(T_i) 
  + \rho\, (\mathbf{A}_i^o)^\top \mathbf{A}_i^o + \notag  \\
  \!& \hspace{-0.5em} \quad \rho \sum_{j=1}^{M} \mathbf{A}^v(T_{ij})^\top \mathbf{A}^v(T_{ij}) \Big]^{-1} \cdot \Big[ 
  \widetilde{\mathbf{M}}(T_i)^\top(\widetilde{z_i} - u_i) + \notag \\
  \!& \hspace{-0.5em} \quad (\mathbf{A}_i^o)^\top(b_i - s_i^k - v_i^k) 
  + \sum_{j=1}^{M} \mathbf{A}^v(T_{ij})^\top(\phi_{ij}^k - w_{ij}^k) \Big], \label{QpSubProblem} \\
  \!& \hspace{-0.5em} z_i^{k+1} = \frac{1}{2} \left( \widetilde{\boldsymbol{\beta}}(T_i) \cdot c_i^{k+1} + \widetilde{\boldsymbol{\beta}}(0) \cdot c_{i+1}^{k+1} \right), \label{UpdateZ} \\
  \!& \hspace{-0.5em} u_i^{k+1} = u_i^k + \widetilde{\mathbf{M}}(T_i) c_i^{k+1} - \widetilde{z_i}^{k+1}, \label{UpdateU} \\
  \!& \hspace{-0.5em} s_i^{k+1} = \max\left(0, -\left( \mathbf{A}_i^o c_i^{k+1} - b_i + v_i^k \right) \right), \label{NonNegativeProject} \\
  \!& \hspace{-0.5em} v_i^{k+1} = v_i^k + \mathbf{A}_i^o c_i^{k+1} + s_i^{k+1} - b_i, \label{UpdateV}\\
  % \!& d_{ij}^{k+1} = \sqrt{ \max\left( \frac{ \left\| \mathbf{A}^v(T_{ij}) c_i^{k+1} \right\|^2 }{ v_{\max}^2 }, 1 \right)^{-1} }, \label{DynamicProject} \\
  \!& \hspace{-0.5em} \phi_{ij}^{k+1} = \sqrt{ \max\left( \left\| \mathbf{A}^v(T_{ij}) c_i^{k+1} + w_{ij}^k \right\|^2 / v_{\max}^2, 1 \right)^{-1} } \notag  \\
  \!& \hspace{-0.5em} \quad \cdot (\mathbf{A}^v(T_{ij}) c_i^{k+1} + w_{ij}^k), \label{DynamicProject} \\
  \!& \hspace{-0.5em} w_{ij}^{k+1} = w_{ij}^k + \mathbf{A}^v(T_{ij}) c_i^{k+1} - \phi_{ij}^{k+1}, \label{UpdateCloseW}
  \end{flalign}  
\end{subequations}
% where $\widetilde{z_i} = [z_{i-1}, z_{i}]^\top$ denotes the components of $z$ associated with the $i$-th segment, and $z_0$ and $z_N$ are initialized by $\overline{\mathbf{\sigma}}_{s}$ and $\overline{\mathbf{\sigma}}_{f}$, respectively, which remain fixed during the iteration. 
% The matrix $\widetilde{M}(t) = [\beta^{[d-1]}(0)^\top, \beta^{[d-1]}(t)^\top]^\top$ maps the coefficients in the $i$-th segment to high-order derivatives at both boundaries.
% To be noticed, each subproblem at each iteration is actually a overdetermined system, where $m \times 2d$ continuity constraints are imposed on objective with $m \times d$ degree of freedom. The penalty team in Augmented Largrangian Formation relaxes the constraints during iteration. 
where the matrix $\widetilde{\boldsymbol{\beta}}(t) = \mathbf{I}_{m\times m} \otimes (\beta^{0}(t), \dots, \beta^{(d-1)}(t))^\top \in \mathbb{R}^{md \times m(d+1)}$.

Note that each subproblem updating $c_i$ at each iteration is an overdetermined system, with $m \times 2d$ continuity constraints imposed on an objective with $m \times (d+1)$ degrees of freedom.
However, a solution exists due to the relaxation of the constraints in the augmented Lagrangian term.
\subsubsection{Solution Existence and Convergence Analysis}
\label{sec:solution_existence_and_convergence}
We guarantee the existence of the solution referring to the \textit{Optimality Conditions}\cite{Gcopter}, which provides the necessary and sufficient conditions for the optimal solution of the minimum control effort problem in Eq. (\ref{minimum_control_effort}) without Eq. (\ref{general_constraints}), as follows:
% the $\sigma(t)$ is optimal for the minimum control effort problem in (\ref{minimum_control_effort}) if the following conditions is satisfied:
\begin{itemize}
  \item The degree of piecewise polynomial $d=2p-1$.
  % in Eq. (\ref{polynomial traj class}).
  \item The boundary conditions are constrained to $(p-1)$-th order derivative in Eq. (\ref{start_points_constraints}).
  \item The continuously differentiable constraints in Eq. (\ref{continously_differentiable_constraints}) are constrained to $(2p-1)$-th order derivative.
\end{itemize}

The only difference in our method is that we do not constrain the intermediate conditions of the \textit{Optimality Conditions}. 
The optimal solution must lie within our solution space, which is larger, thus ensuring that the solution can be found.
% Moreover, The foundation algorithm framework base on the ADMM\cite{Boyd} ensure the objective will convergence to one of the minimum solutions when iterations step k goes to infinity.
% a little long
Moreover, the convexity of the formulated problem and the foundation of the ADMM\cite{Boyd} framework ensures that the objective converges to an optimal solution as the number of iterations \(k\) approaches infinity.
\setlength{\textfloatsep}{2pt}
\begin{algorithm}[!t]
  \caption{\textbf{TOP Algorithm Framework}}  
  \DontPrintSemicolon
  \LinesNumbered
  \SetAlgoLined
  \label{alg_TrajSplit_CADMM_TBB}
  % \textbf{Notion:} 
  % ~number of trajectory segments $N$,\;
  % ~~~~~~~~~~~polynomial coefficients: $c_i$,\;
  % ~~~~~~~~~~~consensus variables: $z_i$,\;
  % ~~~~~~~~~~~dynamic consensus variables: $d_i$,\;
  % ~~~~~~~~~~~scaled dual variables: $u_i$, $v_i$, $w_i$,\;
  % ~~~~~~~~~~~objective function $f$,\;
  % ~~~~~~~~~~~splitting tolerance $r^k$, $d^k$,\;
  % ~~~~~~~~~~~initial penalty parameter $\rho^0$,\;
  % ~~~~~~~~~~~primal/dual residual tolerance $\mathcal{E}_{r}$, $\mathcal{E}_{d}$\;
  % ~~~~~~~~~~~coefficient matrix $C_s$, $A_i$ \;
  \renewcommand{\algorithmicfor}{\textbf{For}}

  \KwIn{path $\mathbf{P}$, flight corridor $\mathcal{P}^{\mathcal{H}}$ \;
  }
  \KwOut{optimized polynomial coefficients $\mathbb{C}$}
  \Begin
  {
      %  \textbf{Init} ($c^0$, $z^0$, $u^0$, $v^0$, $s^0$, $u_i$, $T_i$).\;
      \textbf{Init} ($\mathbf{P}$, $\mathcal{P}^{\mathcal{H}}$)\;
      \While{$\|r_p^k\|^2 < \mathcal{E}_r^2$ and $\|r_d^k\|^2 < \mathcal{E}_d^2$}
      {
          %  \textbf{ParallelFor} ($i=1$ to $N$) \textbf{do}\;
          \textbf{Parallel} \For{$i=1$ to $N$}
          {
              % $\textcolor{blue}{\text{/* update $c_i$ */}}$\;
              % \hspace{1cm}{
              \eIf{closed-form}
              {
                  %  $c_i^{k+1} \leftarrow$ \textbf{QpSubProblem($c_i^k, z_i^k,d_i^k$)}\;
                  %  $d_{ij}^{k+1} \leftarrow$ \textbf{DynamicProject($d_{ij}^k$)}\;
                  % %  $w_{ij}^{k+1} \leftarrow w_{ij}^{k} + {\mathbf{A}^v}(t_{ij})c_i^{k+1} - d_{ij}^{k+1}$\;
                  %  $w_{ij}^{k+1} \leftarrow$ \textbf{UpdateCloseW($w_{ij}^{k}$)}\;
                   $c_i^{k+1}, \phi_{i}^{k+1}, w_{i}^{k+1} \leftarrow$ \textbf{ClosedFormSolve($c_i^k, z_i^k, \phi_i^k, u_i^k, s_i^k, v_i^k, w_i^k$)}
              }
              {
                  %  $c_i^{k+1} \leftarrow$ \textbf{OptSolver($c_i^k, z_i^k$)}\;
                  %  $w_{ij}^{k+1} \leftarrow w_{ij}^{k} + g(c_i^{k+1})$\;
                   $c_i^{k+1}, w_{i}^{k+1} \leftarrow$ \textbf{NumericSolve($c_i^k, z_i^k,u_i^k, s_i^k, v_i^k, w_i^k$)}
              }
          {
              %  $\textcolor{blue}{\text{/* update $z_i$ */}}$\;
              %  $z_i^{k+1} \leftarrow \frac{1}{2} {M_s} (c_{i}^{k+1} + c_{i+1}^{k+1})$\;
               $z_i^{k+1} \leftarrow$ \textbf{UpdateConsensus($c_i^{k+1}$)}\;
               $r_p^{k+1}, r_d^{k+1} \leftarrow $ \textbf{UpdateResidual($c_i^{k+1},z_i^{k+1}$)}\;
                % $d^k \leftarrow $ \textbf{updateDualResidual($z_i$,$z_{i+1}$)} 
          }
          {
              % $\textcolor{blue}{\text{/* update $s_i$ */}}$\;
               $s_i^{k+1} \leftarrow$ \textbf{ProjectNonNeg($c_i^{k+1}, v_i^k$)}
          }\;
          {
              % $\textcolor{blue}{\text{/* update dual values */}}$\;
              %  $u_i^{k+1} \leftarrow u_i^{k} + {M_s}c_i^{k+1} - z_i^{k+1}$\;
              %  $u_i^{k+1} \leftarrow \textbf{UpdateDualU ($u_i^{k}$)}$\;
              %  $v_i^{k+1} \leftarrow v_i^{k} + {A^s_i}c_i^{k+1} + s_i^{k+1} + b_i$\;
              %  $v_i^{k+1} \leftarrow \textbf{UpdateDualV ($v_i^{k}$)}$\;
               $u_i^{k+1},v_i^{k+1} \leftarrow \textbf{UpdateDual ($c_i^{k+1},z_i^{k+1},s_i^{k+1},u_i^{k},v_i^{k}$)}$\;
              %  \eIf {closed-form}
              %  {
              %      $w_{ij}^{k+1} \leftarrow w_{ij}^{k} + {A^v}(t_{ij})c_i^{k+1} - d_{ij}^{k+1}$\;
              %  }
              %  {
              %   %  w^k+1 = w^k + rho*(max{0,||A_vec[iterator][j]*x||^2- v_max_2})^2
              %     % $w_{ij}^{k+1} \leftarrow w_{ij}^{k} + \rho \cdot \max(0, ||A^v(t_{ij})c_i^{k+1}||^2 - v_{max}^2)$\;
              %     $w_{ij}^{k+1} \leftarrow \textbf{GInequalityDualUpdate($c_i^{k+1}$)}$\;
              %  }
          }
          {
              % $\textcolor{blue}{\text{/* update penalty parameter $\rho$ */}}$\;
              $\rho^{k+1} \leftarrow$ \textbf{UpdateRho($\rho^k$)}\;
          }
          }
          %  \textbf{end ParallelFor}\;
          % \EndFor
      }
      % \EndWhile
      \Return{$\mathbb{C}$}\;
  }
\end{algorithm}
\vspace{-0.1cm}
\subsubsection{Stopping Criterion}
The stopping criterion for \textbf{TOP} is determined by the primal and dual residuals.
The primal residual is defined as the gaps between state variables at the splitting points:
\begin{align}
  r_p^{k+1} &= \left( {\widetilde{\boldsymbol{\beta}}(T_1)} c_1^{k+1} - {\widetilde{\boldsymbol{\beta}}(0)} c_{2}^{k+1}; \cdots;\right. \nonumber \\
      &~~~~~~~~\quad \left.  {\widetilde{\boldsymbol{\beta}}(T_{N-1})} c_{N-1}^{k+1} - {\widetilde{\boldsymbol{\beta}}(0)} c_{N}^{k+1} \right).
  \label{splitting_primal_residual}
\end{align}
% The dual residual is defined as the difference between the current Consensus variable and the one in the last iteration.
The dual residual reflects the convergence degree of the dual variables update, defined as:
\begin{align}
  r_d^{k+1} &= -\rho \widetilde{\mathbf{M}}(T_i)^\top \left( \widetilde{z}_1^{k+1} - \widetilde{z}_1^k;  \cdots;\right. \nonumber \\
  &~~~~~~~~\quad \left. \widetilde{z}_{N-1}^{k+1} - \widetilde{z}_{N-1}^k; \widetilde{z}_N^{k+1} - \widetilde{z}_N^k \right).
  \label{splitting_dual_residual}
\end{align}
The algorithm will stop when the primal residual and dual residual are both smaller than the predefined tolerance.
\begin{equation}
  \|r_p^k\|^2 <\mathcal{E}_r^2, ~~\|r_d^k\|^2 < \mathcal{E}_d^2,
  \label{splitting_tolerance}
\end{equation}
where \(\mathcal{E}_r = \mathcal{E}_d = N\mathcal{E}\). In practice, we recommend  \(\mathcal{E} = 0.05\).
vspace{-0.1cm}
\subsubsection{Adaptive Rho}
To speed up the convergence of the algorithm, we update the penalty parameter \(\rho^k\) adaptively, making the optimization less sensitive to the initial value \(\rho^0\). 
The penalty parameter \(\rho\) is updated in each iteration according to the following rule:
\begin{equation}
  \rho^{k+1}=\left\{
  \begin{array}{ll}
    \tau^{\text {incr }} \rho^{k} & \text { if }\left\|r_p^{k}\right\|_{2}>\mu\left\|r_d^{k}\right\|_{2}, \\[5pt]
    \rho^{k} / \tau^{\text {decr }} & \text { if }\left\|r_d^{k}\right\|_{2}>\mu\left\|r_p^{k}\right\|_{2}, \\[4pt]
    \rho^{k} & \text { otherwise },\end{array}\right.
    \label{UpdateRho}
\end{equation}
where $\mu>1$, $\tau^{\text {incr }}>1$, and $\tau^{\text {decr }}>1$. 
In our implementation, we set $\mu = 10$, $\tau^{\text {incr }}=\tau^{\text {decr }}=1.1$, and $\rho^0=1$.
% In experiments, we set $\mu = 10$, $\tau^{\text {incr }}=\tau^{\text {decr }}=1.1$, $\rho^0=1$ and .
% Typically, $\mu = 10$ , $\tau^{\text {incr }}=\tau^{\text {decr }}=2$. 
The strategy aims to maintain the primal and dual residuals in the same order of magnitude to accelerate the optimization process.
\subsubsection{Algorithm Summary}
The framework of \textbf{TOP} is outlined in Alg. \ref{alg_TrajSplit_CADMM_TBB}.  
It takes the path $\mathbf{P} \in \mathbb{R}^{m\times(N+1)}$ found by the RRT* and the flight corridor $\mathcal{P}^{\mathcal{H}}$ generated by the FIRI as inputs. 
Line~2 (\textbf{Init}) sets up all variables for each segment. 
After initialization, all trajectory segments are optimized in parallel. 

% If the closed-form solution is chosen, the \textbf{QpSubProblem} in Line~7 is solved using~(\ref{QpSubProblem}), which meets most quadrotor requirements. 
% Then, the dynamic projection is executed in Line~8 using~(\ref{DynamicProject}), followed by the dual variable update in Line~9. 
If the closed-form solution is chosen, line~6 (\textbf{ClosedFormSolve}) applies Eqs. (\ref{QpSubProblem}, \ref{DynamicProject}, \ref{UpdateCloseW}) to update $c_i, \phi_i, w_i$ in the closed-form.
Constraints that cannot be addressed by the aforementioned method can be solved numerically (\textbf{NumericSolve}) using Eq. (\ref{general_constraint_aug}). 
% Line~12 updates the dual variable for general inequality constraints through \(g(c_i^{k+1})\) using~(\ref{componentwise maximum}). 
Afterward, the consensus variable \(z_i\) is updated in line~10 using Eq. (\ref{UpdateZ}). 
Line~11 (\textbf{UpdateResidual}) updates the primal and dual residuals using Eqs. (\ref{splitting_primal_residual}, \ref{splitting_dual_residual}).  
Line~12 (\textbf{ProjectNonNeg}) projects the slack variable \(s_i\) into the non-negative space using Eq. (\ref{NonNegativeProject}). 
The dual variables \(u_i\) and \(v_i\) are subsequently updated (\textbf{UpdateDual}) using Eqs. (\ref{UpdateU},\ref{UpdateV}) in line~13. 
Finally, line~14 updates the penalty parameter \(\rho\) via Eq. (\ref{UpdateRho}). 
The algorithm terminates once both the primal and dual residuals drop below the predefined tolerance, returning the optimized polynomial coefficients $\mathbb{C} = \{c_1,c_2,\dots,c_N\}$. 

  % we can apply the same processing approach to the $i$th trajectory segment, subject to the safety constraints of the flight corridor. The update rule corresponding to the safety constraints is:

% Line 2 initializes all the variables, mainly the polynomial coefficients \(c_i\) and time allocation \(t_i\) for each segment. 
% If the closed-form solution choosen, the \textbf{QpSubProblem} will be solved in line 7 using (\Ref{QpSubProblem}). 
% Then the dynamic projection is performed in line 8 using (\ref{DynamicProject}) with the dual variable updating in line 9. 
% If the closed-form solution is not available, the \textbf{OptSolver} can be applied in line 11 to solve the subproblem using optimization solver. 
% The $g(c_i^{k+1})$ in line 12 is the dual variable updating for the general inequality constraints using (\ref{componentwise maximum}).
% After that, the consensus variable \(z_i\) is updated in line 15. 
% The \textbf{UpdateResidual} in line 16 is used to update the primal and dual residuals using (\ref{splitting_primal_residual}) and (\ref{splitting_dual_residual}). 
% The \textbf{NonNegativeProject} in line 18 is used to project the slack variable \(s_i\) to the non-negative space using (\ref{NonNegativeProject}). 
% The dual variables \(u_i\) and \(v_i\) are updated in line 20 and line 21, respectively. 
% Finally, the penalty parameter \(\rho\) is updated in line 24 using (\ref{UpdateRho}). 

\begin{figure*}[!ht]
  \centering
  \includegraphics[width=0.95\textwidth]{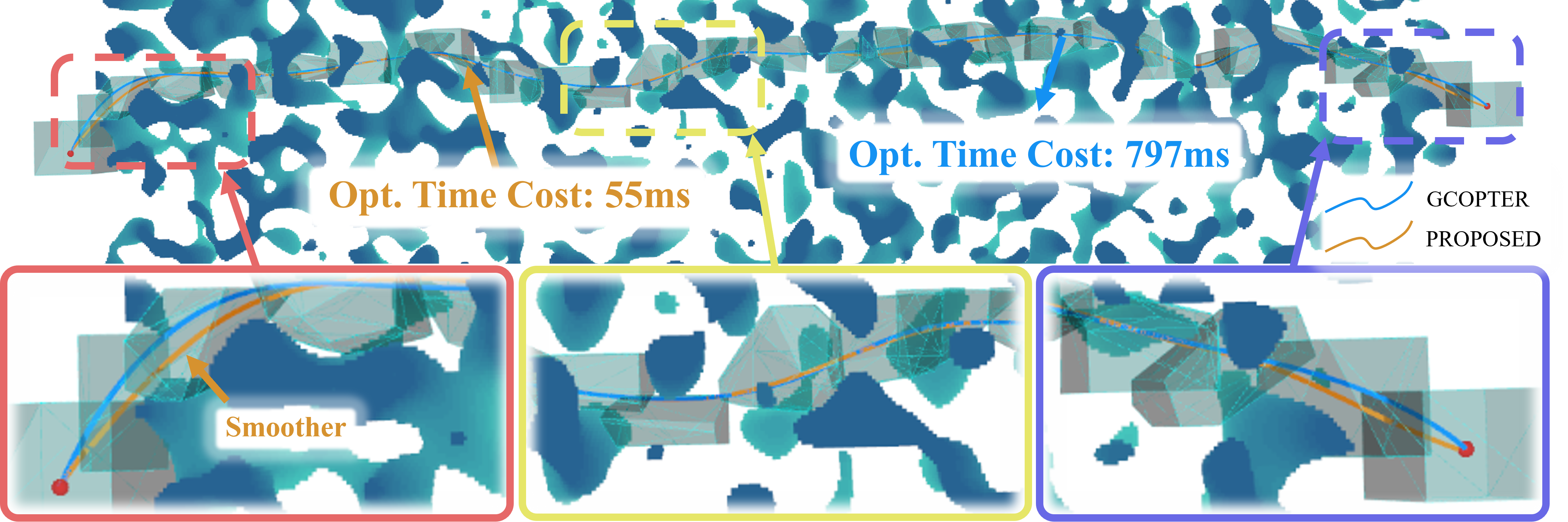}
  \caption{
      Benchmark on the trajectory optimization in simulation. 
  }
  \label{fig: Benchmark}
  \vspace{-0.3cm}
\end{figure*}

% \subfile{experiments.tex}
\vspace{-0.2cm}
\section{Experiments}
\label{Sec: Experiments}
In this section, we conduct a series of experiments with a quadrotor as the target platform to evaluate the performance of the proposed method.
% In all experiments, we choose the quadrotor as the hardware platform, which has been widely researched in the past decades.
The flat output of the quadrotor is defined as $\sigma(t) = (p_x(t), p_y(t), p_z(t), \psi(t))^\top$, where $(p_x(t), p_y(t), p_z(t))^\top$ is the position of the quadrotor in the world frame, and $\psi(t)$ is yaw angle of the quadrotor.
Similar to GCOPTER\cite{Gcopter}, we optimize the position of the quadrotor in experiments while fixing the yaw angle during optimization.

%%%%%%%%%%%%%%%%%%%%%%%%%%%%%%%%%%%%%%%%%%%%%%%%%%%%%%%%%%%%%%%%%%%%%%%%%%%%%%%%
%%%%%%%%%%%%%%%%%%%%%%%%%%%%%%%%%%%%%%%%%%%%%%%%%%%%%%%%%%%%%%%%%%%%%%%%%%%%%%%%
%%%%%%%%%%%%%%%%%%%%%%%%%%%%%%%%%%%%%%%%%%%%%%%%%%%%%%%%%%%%%%%%%%%%%%%%%%%%%%%%
\vspace{-0.30cm}
\subsection{Simulations}
\label{Sec: Simulations}
% In simulations, we first compare the performance with the state-of-the-art method in Sec.\ref{Sec: Benchmark}.
% Next, we analysis the performance of the proposed method under different conditions in Sec.\ref{Sec: Ablation}. 
%%%%%%%%%%%%%%%%%%%%%%%%%%%%%%%%%%%%%%%%%%%%%%%%%%%%%%%%%%%%%%%%%%%%%%%%%%%%%%%%
\subsubsection{Benchmark}
\label{Sec: Benchmark}
\begin{table}[h]
  \captionsetup{font={small}}
  \caption{
    Benchmark on the Trajectory Optimization
    }
  \label{tab:Benchmark}
  \centering
  \renewcommand{\arraystretch}{1.3}
  \Large
  \resizebox{\columnwidth}{!}{
      \begin{tabular}{cccccc}
          \toprule[1pt]
          \multirow{2}{*}{Piece Num} & \multirow{2}{*}{Method} & \multicolumn{2}{c}{Obj. Cost} & \multicolumn{2}{c}{Time Cost ($ms$)} \\
          & & Avg & Std & Avg & Std \\
          % \midrule
          % % \multirow{3}{*}{10$\sim$19} 
          % %     & Proposed (Parallel) & 29.6953 & 20.1893 & 104.8795 & 63.4356 \\
          % %     & Proposed (Sequential)  & 29.6953 & 20.1893 & 249.2707 & 143.1241 \\
          % %     & GCopter & 96.7815 & 19.5109 & 99.8 & 71.2752 \\
          % % \midrule
          % \multirow{3}{*}{20$\sim$29} 
          %     & Proposed (Parallel) & \textbf{39.8079} & \textbf{18.2108} & \textbf{85.6315} & \textbf{23.2205} \\ba
          %     & Proposed (Sequential) & 39.8079 & 18.2108 & 215.8356 & 67.9761 \\
          %     & GCopter & 130.3441 & 22.6431 & 184.6 & 94.2955 \\
          \midrule
          \multirow{3}{*}{30$\sim$39} 
              & Proposed (Parallel) & \textbf{34.1336} & \textbf{26.0209} & \textbf{78.2922} & \textbf{17.6207} \\
              & Proposed (Sequential) & 34.1336 & 26.0209 & 235.1034 & 60.2844 \\
              & GCOPTER & 179.6086 & 42.0543 & 557.3 & 181.9148 \\
          \midrule
          \multirow{3}{*}{40$\sim$49} 
              & Proposed (Parallel) & \textbf{30.9601} & \textbf{12.3711} & \textbf{88.9753} & \textbf{33.9521} \\
              & Proposed (Sequential) & 30.9601 & 12.3711 & 315.6224 & 139.2553 \\
              & GCOPTER & 232.0403 & 20.2153 & 745.7 & 242.4145 \\
          \midrule
          \multirow{3}{*}{50$\sim$59} 
              & Proposed (Parallel) & \textbf{60.9266} & \textbf{26.8216} & \textbf{108.864} & \textbf{25.8078} \\
              & Proposed (Sequential) & 60.9266 & 26.8216 & 485.6633 & 140.6211 \\                                                              
              & GCOPTER & 296.7563 & 28.1774 & 877.0 & 527.0681 \\
          \midrule
          \multirow{3}{*}{60-69} & 
          Proposed(Parallel) & \textbf{64.3979} & \textbf{27.6639} & \textbf{118.397} & \textbf{22.0706} \\
          &Proposed(Sequential) & 64.3979 & 27.6639 & 564.6306 & 120.1771 \\
          &GCOPTER & 297.1527 & 58.2767 & 1288.8 & 407.8276 \\
          \toprule[1pt]
      \end{tabular}
  }
\end{table}
% All methods evaluated in this experiment is implemented in C++.
In this experiment, all methods are implemented in C++ for performance.
We adopt Thread Building Blocks (TBB) to parallelize the computation of the proposed method on the CPU.
The experiment is conducted on a computer with an AMD Ryzen Threaddripper Pro 5995wx CPU.
We compare the performance of our method with GCOPTER in two dimensions: optimized objective cost and optimization time cost.
The \textbf{Proposed(Parallel)} and \textbf{Proposed(Sequential)} indicate the proposed method with and without parallelization respectively.
The optimized trajectories are visualized in Fig.\ref{fig: Benchmark}, where the trajectory of GCOPTER exhibits higher curvature.
The results listed in Tab. \ref{tab:Benchmark} demonstrate that our method outperforms GCOPTER in both smoothness and time cost. 
% Note that the parameters of GCOPTER are modified in this experiment to ensure both methods are compared under same constraints and initial values.
Note that both methods are ensured to have the same constraints and initial values for fairness.
In GCOPTER, the original constrained problem is transformed into an unconstrained optimization problem and solved using gradient descent method. 
% In AllocNet, it uses a neural network to give the time allocation for trajectories and leverages a QP solver to solve the quadratic problem.
However, its architecture does not effectively support high-quality parallelization, resulting in high time cost as the problem scale increases.
% Our proposed method has a better performance than GCOPTER in terms of both objective cost and time cost.
% Unfortunately, we indeed found that CPU may not that suitable for parallel optimization.
Note that the time cost of our method also increases slightly with the number of segments, which does not show the constant time complexity.
% The reason lies in CPU is scheduled by the operating system kernel to perform tasks within the computer, which is not as specific as GPU. 
The reason is that the CPU, scheduled by the operating system kernel to handle general-purpose tasks, is not as specialized for parallelization.
Meanwhile, there are usually far fewer cores in a CPU than in a GPU, which also limits the parallelization capability.
Therefore, in the next section, we conduct experiments on a GPU to validate the parallelization performance of our method.
% That's the reason why proposed method does not perform ideal parallelization in this experiment.

\subsubsection{Ablation Study}
\label{Sec: Ablation}

In this experiment, we evaluate the effectiveness of the \textbf{UpdateRho} function in Alg.~\ref{alg_TrajSplit_CADMM_TBB} (line~14) by comparing the performance of the proposed method with several variations. 
% The trajectory to be optimized has 56 segments and travels from $(-119.542, -119.941, 0.5)^\top$ to $(119.036, 119.691, 0.5)^\top$.
% We first remove the adaptive $\rho$ method and fix the $\rho$ to different values (1, 10, and 100 in experiment).
Specifically, we remove the \textbf{UpdateRho} function and fix \(\rho\) to different values (1, 10, and 100) as control groups.
The \textbf{UpdateRho} dynamically adjusts \(\rho\) during optimization to balance the convergence of the primal and dual residuals, allowing our proposed method to achieve faster convergence compared to the fixed \(\rho\) method.
The details of these comparisons are presented in Tab. \ref{tab: Ablation Study}.
The hardware setup remains consistent with the experiment described in Sec.~\ref{Sec: Benchmark}.
% We also compare the performance of the proposed method with and without the LBFGS method in update x\textbf{TODO: change to the more formal representation}.
% The LBFGS method is a quasi-Newton method that approximates the Hessian matrix of the objective function.
% The results show that our method consumes similar iterations with using the LBFGS during optimization.
% However, parallelizing LBFGS is more expensive than our proposed method, which is not suitable for real-time applications.
% Overall, the proposed method has a better convergence speed and lower time cost than the other methods.
\vspace{-0.1cm}
\begin{table}[h]
    \caption{
        Ablation Study on Updating $\rho$
    }
    \label{tab: Ablation Study}
    \centering
    % \normalsize
    \renewcommand{\arraystretch}{1.3}
    \Large
    \resizebox{\columnwidth}{!}
    {
        \begin{tabular}{cccccc}
            \toprule[1.5pt]
            Method & Iter. Num. & Pri. Res. & Dual Resi. & Obj. Cost & Time Cost ($ms$) \\
            \midrule
            Adaptive $\rho$ & \textbf{416} & 0.7721 & \textbf{0.1817} & \textbf{66.8705} & \textbf{149.1764} \\
            Fixed $\rho$ (1) & 607 & 1.9738$e^{-2}$ & 0.7696 & 228.6242 & 195.4008 \\
            Fixed $\rho$ (10) & 5120 & 1.3566$e^{-3}$ & 0.7741 & 282.6921 & 1598.8727\\
            Fixed $\rho$ (100) & 23778 & \textbf{9.0375$e^{-13}$} & 0.7741 & 273.4577 & 7440.5748\\
            % With LBFGS & \textbf{267} & 0.7727 & \textbf{0.1569} & \textbf{46.4666} & 1490.0212\\
            \toprule[1.5pt] 
        \end{tabular}
    }
\end{table}
\vspace{-0.5cm}
\begin{table}[h]
  \caption{
      Comparison Between Closed-form Method and Numerical Method
  }
  \label{tab: Close Form vs Numerical}
  \centering
  \fontsize{5pt}{6pt}\selectfont % 设置字体大小为10pt，行间距为12pt
  \resizebox{\columnwidth}{!}
  {
      \begin{tabular}{cccc}
          \toprule[0.5pt]
          Method & Mean Iter. Num. & Mean Time Cost (ms) \\
          \midrule
          Closed-form & {230.9} & \textbf{88.6334}  \\
          Numerical & \textbf{157.2} & 1169.7684 \\
          \toprule[0.5pt] 
      \end{tabular}
  }
  \vspace{-0.3cm}
\end{table}

% Last, we compare the performance of the proposed method in adopting the closed-form method and the numerical method in update $\mathbf{x}$.
Next, we compare the performance of the proposed method using both the closed-form and numerical methods, as outlined in Alg. \ref{alg_TrajSplit_CADMM_TBB} (line~5-9).
We also reuse the same hardware setting as the experiment in Sec. \ref{Sec: Benchmark}.
The results are listed in Tab.\ref{tab: Close Form vs Numerical}. 
We find that the general constraints formulation allows the numerical method to converge after fewer iterations. 
However, the closed-form method is significantly more time-efficient than the numerical method, making it better suited for real-time applications.
\begin{figure}[h]
  \centering
  \vspace{-0.35cm}
  \includegraphics[width=0.476\textwidth]{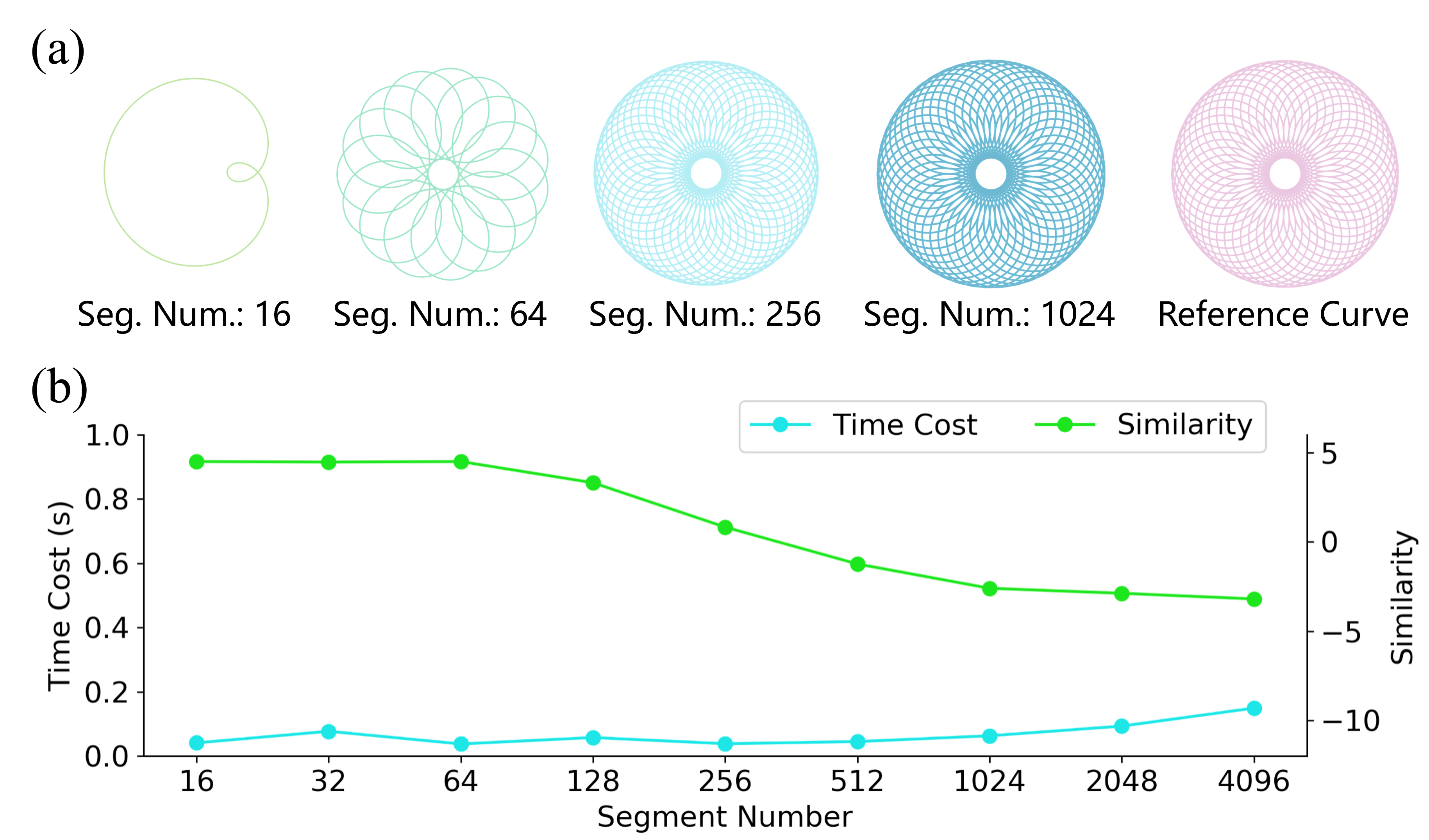}
  \caption{(a) visualizes the growth of the similarity with the segment number.
           (b) demonstrates the relationship between computation time, similarity, and segment number.}
  \label{fig:exp_sim}
\end{figure}
\vspace{-0.2cm}

Finally, we compare the performance of our approach with different segment numbers.
We uniformly sample $N+1$ initial points from a parameterized curve and enforce the optimized trajectories to pass through the sampled points, forming \(N\) segments. 
In this experiment, we implement our code in Python and leverage GPU for parallelization.
% The hardware used in this experiment is \textbf{TODO: Hardware}.
The computation is conducted on a computer with a NVIDIA GeForce RTX 4060 Laptop GPU. 
In this experiment, we compare not only the computation time of optimization under different segment numbers but also evaluate the similarity of the optimized trajectories to the reference curve. 
The metric is defined as $e_{simi} = lg(\sum_{i=1}^M |{p}_i - {p}^{ref}_{ i}|),$
where \(M\) denotes the number of uniformly sampled points used for evaluation, set to \(8192\) in practice, and \({p}_i\) and \({p}^{ref}_{i}\) represent the \(i\)-th point of the optimized trajectory and the reference curve, respectively. 
% To avoid the influence of the initial values, we perform 3000 iterations in optimization under all conditions. 
The results in Fig.~\ref{fig:exp_sim} demonstrate the potential of our method, as it maintains a similar runtime even as the number of segments increases, confirming its parallelization capability.
Meanwhile, the similarity of the optimized trajectories decreases as the segment number grows, which illustrates the nonnegligible of the segment number when a precise trajectory is required.
\begin{figure*}[!h]
  \centering
  \raisebox{0cm}{\hspace{0.2cm}
  \includegraphics[width=0.98\textwidth]{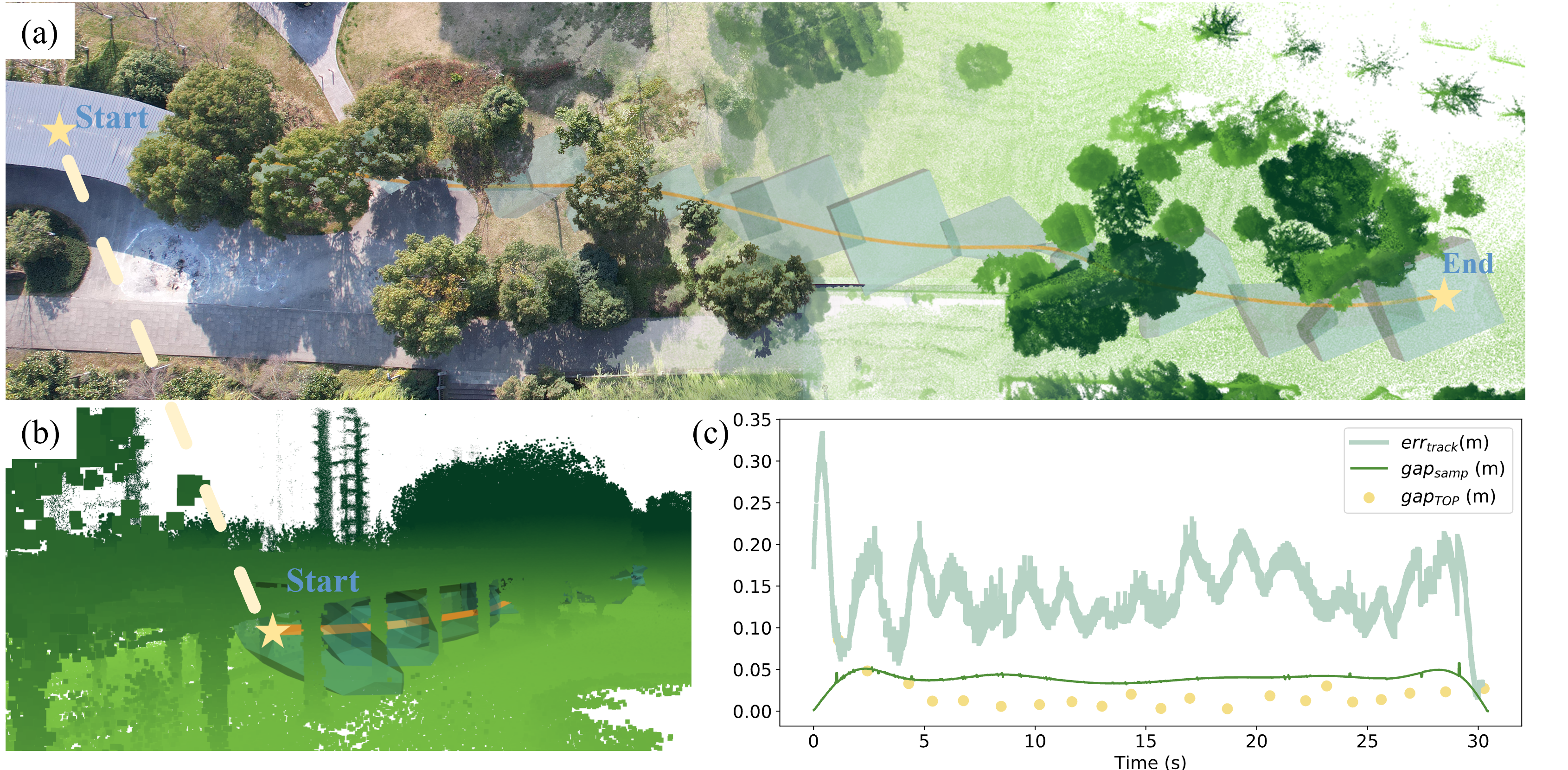} % 调整图片宽度为页面宽度
  }
  \caption{
    The top-down (a) and side (b) views of the optimized trajectory and flight corridor in a real-world environment and point cloud map.
    % In (c), the tracking error and the position gap between pieces of the optimized trajectory are visualized, where the gap reaches an acceptable level that far less than the tracking error. 
    In (c), as the $gap_{TOP}$ is lower than the $gap_{samp}$ in most cases and there are no sudden changes in \(err_{track}\) when encountering \(gap_{TOP}\), we argue that the gap $gap_{TOP}$ would not influence the performance of the quadrotor substantially. 
    }
    \vspace{-0.65cm}
  \label{fig:real_world}
\end{figure*}

\vspace{-0.4cm}
\subsection{Real-world Tests}
\label{Sec: Real-world}

To validate the performance of the proposed method in real-world applications, we conduct a series of experiments with a customized quadrotor platform.
The quadrotor is equipped with a NVIDIA Orin NX for on-board computation, and a Livox MID360 for perception.
We adopt a modified version of FAST-LIO2 \cite{FastLio2} with a relocalization module for the robot's state estimation. 

\subsubsection{Traverse in Forest}
\label{Sec: Traverse in Forest}

In this experiment, we test the performance of the proposed method in a forest environment, where the quadrotor is required to traverse through trees and obstacles. 
The point cloud map is built offline, then we relocalize the quadrotor's pose in the map for planning.
% The quadrotor is required to follow a trajectory that passes through the trees while avoiding collisions with them.
We use RRT* to generate a path and then apply the proposed method to optimize the trajectory, producing a smooth and agile trajectory with a maximum velocity of \(5 \, \text{m/s}\). 
The result is shown in Fig. \ref{fig:real_world}, where the quadrotor successfully traverses through the trees without collisions.
% Meanwhile, the position gaps between pieces of trajectory \textbf{cite the figure} reaches at an acceptatble level that much lower than the tracking error of the quadrotor.
To demonstrate that the position gap between trajectory segments ($gap_{TOP}$) is acceptable, we compare it with the position gap of sampled discretized state $gap_{samp}$ and the tracking error \(err_{track}\)(Fig. \ref{fig:real_world}(c)). 
% First, as the controller tracks the discretized states (position, velocity, etc.) on the trajectory, we found in most cases, the position gap of the discretized states $gap_{samp}$ is higher than $gap_{TOP}$, which indicates that $gap_{TOP}$ would not take the main part.
First, we observe the $gap_{TOP}$ is consistently smaller than the $gap_{samp}$ along the trajectory, indicating that it does not impact the controller's ability to track the trajectory.
%This indicates that $gap_{TOP}$ has a minor impact on the overall performance. 
Second, the fluctuations in the \(err_{track}\) show no evidence of being correlated with \(gap_{TOP}\), as observed in Fig.~\ref{fig:real_world}(c). 
Furthermore, around 15 centimeters position's tracking errors $err_{track}$ are acceptable considering \(5 \, \text{m/s}\) speed flying.
Therefore, we prove the $gap_{TOP}$ between segments on the trajectory is acceptable for quadrotor control in real-world tests.

\subsubsection{Paint Complex Geometry in the Sky}
\label{Sec: Track}

In this experiment, we generate a $538.14 m$ dynamical feasible trajectory with complex geometry via the proposed method.
The trajectory consists of 1000 segments and requires only 0.037 seconds for computation.
The Fig. \ref{fig:real_gpu_test} demonstrates that our method produces high-quality trajectories that are friendly for the quadrotors.
Note that the complex geometry of the trajectory requires numerous segments to recover the paint, highlighting the advantages of parallelism in our method.
\vspace{-0.3cm}
\section{Conclusion}
In this paper, we propose \textbf{TOP}, a trajectory optimization method via parallel optimization towards constant time complexity. 
% \todo{not title, in content, so no need to capitalize all the letter}. 
% TOP: Trajectory Optimization via Parallel Optimization towards Constant Time Complexity 
The proposed method is based on the CADMM framework, enabling parallel optimization of multiple trajectory segments and reducing the time complexity to a constant level. 
Experimental results show that the proposed method is tenfold times faster than the SOTA approach when optimizing a large-scale trajectory with one hundred segments, while maintaining superior trajectory quality. 
Additionally, we introduce two approaches to solve the subproblem: a closed-form solution and a numerical solution. 
The former is more time-efficient, while the latter is more general and can handle various constraints. 
In the future, we will explore adaptive resplitting and remerging of trajectories within the environment to further enhance the efficiency of the proposed method.
\vspace{-0.2cm}

\bibliographystyle{IEEEtran}
\bibliography{paper}

\end{document}